\let\lineseg\line
\newenvironment{numbered}{\medskip\par\noindent
  \begin{equation}\small\begin{minipage}{.9\textwidth}}{%
  \end{minipage}\end{equation}}
\documentstyle[twoside,cl]{article}
\issue{Vol}{Num}{199X}
\title{An Efficient Implementation of the Head-Corner Parser}
\author{Gertjan van Noord%
  \thanks{Alfa-informatica \& BCN,
         \tt vannoord@let.rug.nl}\\
      \affil{Rijksuniversiteit Groningen}}
\runningtitle{Efficient Head-Corner Parsing}
\runningauthor{van Noord}

\begin{document}

\maketitle

\begin{abstract}
  This paper describes an efficient and robust implementation of a
  bi-directional, head-driven parser for constraint-based grammars.
  This parser is developed for the OVIS system: a Dutch spoken dialogue
  system in which information about public transport can be obtained
  by telephone. 

  After a review of the motivation for head-driven parsing strategies,
  and head-corner parsing in particular, a non-deterministic version
  of the head-corner parser is presented.  A memoization technique is
  applied to obtain a fast parser.  A {\em goal-weakening} technique
  is introduced which greatly improves average case efficiency, both
  in terms of speed and space requirements.

  I argue in favor of such a memoization strategy with goal-weakening
  in comparison with ordinary chart-parsers because such a strategy
  can be applied selectively and therefore enormously reduces the
  space requirements of the parser, while no practical loss in
  time-efficiency is observed. On the contrary, experiments
  are described in which head-corner and left-corner parsers
  implemented with selective memoization and goal weakening outperform
  `standard' chart parsers. The experiments include the grammar of
  the OVIS system and the Alvey NL Tools grammar.

  Head-corner parsing is a mix of bottom-up and top-down processing.
  Certain approaches towards robust parsing require purely bottom-up
  processing. Therefore, it seems that head-corner parsing is
  unsuitable for such robust parsing techniques.  However, it is shown
  how underspecification (which arises very naturally in a logic
  programming environment) can be used in the head-corner parser to
  allow such robust parsing techniques.  A particular robust parsing
  model is described which is implemented in OVIS.
\end{abstract}

\section{Motivation}

In this paper I discuss in full detail the implementation of the
head-corner parser. But first I describe the
motivations for this approach. I will start by considerations that
lead to the choice of a head-driven parser; I will then argue why
Prolog is an appropriate language for the implementation of the
head-corner parser. 

\subsection{Head-driven Processing}

Lexicalist grammar formalisms, such as Head-driven Phrase Structure
Grammar (HPSG), have two characteristic properties. Lexical elements
and phrases are associated with categories that have considerable
internal structure. Second, instead of construction specific rules, a
small set of generic rule schemata is used. Consequently, the set of
constituent structures defined by a grammar cannot be `read off' the
rule set directly, but is defined by the interaction of the rule
schemata and the lexical categories.  Applying standard parsing
algorithms to such grammars is unsatisfactory for a number of reasons.
Earley parsing is intractable in general, as the rule set is simply
too general. For some grammars, naive top-down prediction may even
fail to terminate.  \namecite{shieber-restriction} therefore proposes a
modified version of the Earley-parser, using {\em restricted} top-down
prediction. While this modification often leads to better termination
properties of the parsing method, in practice it easily leads to a
complete trivialization of the top-down prediction step, thus leading
to inferior performance.

Bottom-up parsing is far more attractive for lexicalist formalisms, as
it is driven by the syntactic information associated with lexical
elements.  Certain inadequacies remain, however. Most importantly, the
selection of rules to be considered for application may not be very
efficient.  Consider, for instance, the following DCG rule:
\begin{numbered}\label{pra}\begin{verbatim}
s([],Sem) --> Arg, vp([Arg],Sem).
\end{verbatim}\end{numbered}
A parser in which application of a rule is driven by the left-most
daughter, as it is for instance in a standard bottom-up active chart
parser, will consider the application of this rule each time an
arbitrary constituent {\tt Arg} is derived. For a bottom-up active
chart parser, for instance, this may lead to the introduction of large
numbers of active items. Most of these items will be useless. For
instance, if a determiner is derived, there is no need to invoke the
rule, as there are simply no {\sc vp}'s selecting a
determiner as subject.  Parsers in which the application of a rule is
driven by the rightmost daughter, such as shift-reduce and inactive
bottom-up chart parsers, encounter a similar problem for rules such as:
\begin{numbered}\label{prb}\begin{verbatim}
vp(As,Sem) --> vp([Arg|As],Sem), Arg.
\end{verbatim}\end{numbered}
Each time an arbitrary constituent {\tt Arg} is derived, the parser
will consider applying this rule, and a search for a matching
{\sc vp}-constituent will be carried out. Again, in many cases (if
{\tt Arg} is instantiated as a determiner or preposition, for
instance) this search is doomed to fail, as a {\sc vp} subcategorizing
for a category {\tt Arg} may simply not be derivable by the grammar.
The problem may seem less acute than that posed by uninstantiated
left-most daughters for an active chart parser, as only a search of
the chart is carried out and no additional items are added to it. 
Note, however, that the amount of search required may grow
exponentially, if more than one uninstantiated daughter is present:
\begin{numbered}\begin{verbatim}
vp(As) --> vp([A1,A2|As]), A1, A2.
\end{verbatim}\end{numbered}
or if the number of daughters is not specified by the rule:
\begin{numbered}\begin{verbatim}
vp([A0]) --> vp([A0,...,An]), A1,..., An.
\end{verbatim}\end{numbered}
as appears to be the case for some of the rule-schemata
used in HPSG.

Several authors have suggested parsing algorithms that may be
more suitable for lexicalist grammars. \namecite{kay-hd} discusses the
concept of {\em head-driven} parsing. The key idea 
is that the linguistic concept {\em head} can be used to
obtain parsing algorithms which are better suited for typical natural
language grammars.  Most linguistic theories assume that among the
daughters introduced by a rule  there is one daughter
which can be identified as the {\em head} of that rule. There are
several criteria for deciding which daughter is the head. Two of these
criteria seem relevant for parsing.  First of all, the head of a rule
determines to a large extent what other daughters may or must be
present, as the head selects the other daughters.
Second, the syntactic category and morphological properties of the
mother node are, in the default case, identical to the category and
morphological properties of the head daughter. These two properties
suggest that it may be possible to design a parsing strategy in
which one first identifies a potential head of a rule, before starting
to parse the non-head daughters. By starting with the head, important
information about the remaining daughters is obtained.  Furthermore,
since the head is to a large extent identical to the mother category,
effective top-down identification of a potential head should be
possible. 

In \namecite{kay-hd} two different head-driven parsers are presented.
First, a `head-driven' shift-reduce parser is presented which differs
from a standard shift-reduce parser in that it considers the
application of a rule (i.e. a reduce step) only if a category matching
the head of the rule has been found. Furthermore, it may shift
elements onto the parse-stack which are similar to the
active items (or `dotted rules') of active chart parsers. By using the
head of a rule to determine whether a rule is applicable, the
head-driven shift-reduce parser avoids the disadvantages of parsers in
which either the leftmost or rightmost daughter is used to drive the
selection of rules.  Kay also presents a `head-corner' parser. The
striking property of this parser is that it does not parse a phrase
from left to right, but instead operates `bidirectionally'.  It starts
by locating a potential head of the phrase and then proceeds by
parsing the daughters to the left and the right of the head.  Again,
this strategy avoids the disadvantages of parsers in which rule
selection is uniformly driven by either the leftmost or rightmost
daughter. Furthermore, by selecting potential heads on the basis of a
`head-corner table' (comparable to the left-corner table of a
left-corner parser) it may use top-down filtering to minimize the
search-space. This head-corner parser generalizes the left-corner
parser. Kay's presentation is reminiscent of the left-corner parser as
presented by \namecite{pereira-shieber} which itself is a
version without memoization of the BUP parser \cite{bup}.

Head-corner parsing has also been considered elsewhere.  In
\namecite{satta-stock}, \namecite{sikkel}, \namecite{sikkel-durbuy}
and \namecite{sikkel-diss} chart-based head-corner parsing for
context-free grammar is considered. It is shown that, in spite of the
fact that bidirectional parsing seemingly leads to more overhead than
left-to-right parsing, the worst-case complexity of a head-corner
parser does not exceed that of an Earley parser.  Some further
variations are discussed in \namecite{markjan-satta}.

In \namecite{head-corner} and \namecite{vannoord-diss} I argue that
head-corner parsing is especially useful for parsing with
non-concatenative grammar formalisms. In \namecite{satta-berlin} and
\namecite{vannoord-tag} head-driven parsing strategies for Lexicalized
Tree Adjoining Grammars are presented.

The head-corner parser is closely related to the {\em
  semantic-head-driven} generation algorithm (cf.\ \namecite{cl} and
references cited there), especially in its purely bottom-up
incarnation.

\subsection{Selective memoization}
The head-corner parser is in many respects different from traditional
chart parsers. An important difference follows from the fact that in
the head-corner parser only larger chunks of computation are
memoized. Backtracking still plays an important role for the
implementation of search.

This may come as a surprise at first. Common wisdom is that although
small grammars may be succesfully treated with a backtracking parser,
larger grammars for natural languages always require the use of a
datastructure such as a {\em chart} or a table of {\em items} to make
sure that each computation is only performed once.  In the case of
constraint-based grammars, however, the cost associated with
maintaining such a chart should not be under-estimated. The memory
requirements for an implementation of the Earley parser for a
constraint-based grammar are often outrageous. Similarly, in an Earley
deduction system too much effort may be spent on small portions of
computation which are inexpensive to (re-)compute anyway. 

For this reason I will argue for an implementation of the head-corner
parser in which only large chunks of computation are memoized. In
linguistic terms, I will argue for a model in which only {\em maximal
  projections} are memoized. The computation that is
carried out in order to obtain such a `chunk' uses a depth-first
backtrack search procedure. This solution dramatically improves upon
the (average case) memory requirements of a parser; moreover it
{\em also} leads to an increase in (average case) time efficiency,
especially in combination with {\em goal-weakening}, because of the
reduced overhead associated with the administration of the chart.  In
each of the experiments discussed in section~\ref{compar} the use of
selective memoization with goal weakening out-performs standard
chart-parsers. 

\subsection{Why Prolog}

Prolog is a particularly useful language for the implementation of a
head-corner parser for constraint-based grammars. This is due to the
following:

\begin{itemize}
\item Prolog provides a built-in unification operation.
\item Prolog provides a built-in backtrack search procedure; 
memoization can be applied selectively. 
\item Underspecification can be exploited to obtain results required
  by certain techniques for robust parsing.
\item Prolog is a high-level language; this enables the application
  of partial evaluation techniques. 
\end{itemize}
These considerations are discussed in turn:

The first consideration does not deserve much further attention. We
want to exploit the fact that the primary datastructures of
constraint-based grammars and the corresponding information-combining
operation can be modelled by Prolog's first order terms and
unification.

As was argued above, Prolog backtracking is {\em not} used to simulate
an iterative procedure to build up a chart via side-effects.  On the
contrary, Prolog backtracking is used truly for search. Of course, in
order to make this approach feasible, certain well-chosen search-goals
are memoized. This is clean and logically well-defined (consider, for
example, \namecite{ds-warren}), even if our implementation in Prolog
uses extra-logical predicates.

The third consideration is relevant only if we are interested in
robust parsing. In certain methods in robust parsing we are interested
in the partial results obtained by the parser. In order to make sure
that a parser is complete with respect to such partial results, it is
often assumed that a parser must be applied that works exclusively
bottom-up. In section~\ref{robustness} it will be shown that the head-corner
parser (which uses a mixture of bottom-up and top-down processing) can
be applied in a similar fashion by using underspecification in the
top-goal. Clearly,
underspecification is a concept that arises naturally in Prolog.

The fact that Prolog is a high-level language has a number of
practical advantages related to the speed of development. A
further advantage is obtained because techniques such as {\em partial
  evaluation} can be applied. For example, I have succesfully applied
the Mixtus partial evaluator \cite{mixtus} to the head-corner parser
discussed below, to obtain an additional 20\% speed increase. In
languages such as C partial evaluation does not seem to be possible
because the low-levelness of the language makes it impossible to
recognize the concepts that are required.

\subsection{Left-corner parsing and head-corner parsing}
As the names suggest, there are many parallels between left-corner and
head-corner parsing. In fact, head-corner parsing is a generalization
of left-corner parsing. Many of the techniques that will be described
in the following sections can be applied to a left-corner parser as
well.  

A head-corner parser for a grammar in which for each rule the
left-most daughter is considered to be the head, will effectively
function as a left-corner parser. In such cases the head-corner parser
can be said to run in `left-corner mode'. Of course, in a left-corner
parser certain simplifications are possible. Based on the experiments
discussed in section~\ref{compar}, it can be concluded that a
specialized left-corner parser is only about 10\% faster than a
head-corner parser running in left-corner mode.  This is an
interesting result: it implies that if a head-corner parser is used,
you can do at least (almost) as good as a left-corner parser, and, as
some of the experiments indicate, often better.

\subsection{Practical relevance of head-corner parsing: efficiency and
  robustness}
The head-corner parser is one of the parsers that is being developed
as part of the NWO Priority Programme on Language and Speech
Technology. An overview of the Programme can be found in 
\namecite{ovisplan}.  An important goal of the Programme is
the implementation of a spoken dialogue system
for public transport information (the OVIS system). The language of
the system is Dutch. 

In the context of the OVIS system, it is important that the parser can
deal with input from the speech recognizer. The interface between the
speech recognizer and the parser consists of word-graphs. In
section~\ref{wordgraphs} I show how the head-corner parser is
generalized to deal with word-graphs.

Moreover, the nature of the application also dictates that the parser
proceeds in a {\em robust} way. In section~\ref{robustness} I discuss
the OVIS Robustness component, and I show that the use of a parser
which includes top-down prediction is not an obstacle towards
robustness.

In section~\ref{compar} we compare the head-corner parser with the
other parsers implemented in the Programme for the OVIS application.
It will be shown that the head-corner parser operates {\em much}
faster than implementations of a bottom-up Earley parser and related
chart-based parsers. Moreover, the space requirements are far more
modest too. The difference with a left-corner parser, which was
derived from the head-corner parser, is small.

We performed similar experiments for the Alvey NL Tools grammar of English
\cite{anlt-grammar}, and the English grammar of the MiMo2 system
\cite{mimo2-article}. From these experiments it can be concluded that
selective memoization with goal-weakening (as applied to head-corner
and left-corner parsing) is substantially more efficient than
conventional chart-parsing. We conclude that at least for some grammars,
head-corner parsing is a good option.

\section{A specification of the Head-corner Parser}
\begin{figure}[tb]
\setlength{\unitlength}{.6pt}

\begin{picture}(120,150)(-150,0)
 \multiput(0,0)(5,10){12}{\makebox(0,0){.}}
 \multiput(0,0)(10,0){12}{\makebox(0,0){.}}
 \multiput(120,0)(-5,10){12}{\makebox(0,0){.}}
 \put(60,125){\makebox(0,0){\tt goal}}
 \put(20,70){\makebox(0,0){1}}
 \put(60,-10){\makebox(0,0){\tt lex}}
 \put(50,0){\line(1,2){10}}
 \put(50,0){\line(1,0){20}}
 \put(70,0){\line(-1,2){10}}
\end{picture}
\begin{picture}(120,150)(-200,0)
 \multiput(0,0)(5,10){12}{\makebox(0,0){.}}
 \multiput(0,0)(10,0){12}{\makebox(0,0){.}}
 \multiput(120,0)(-5,10){12}{\makebox(0,0){.}}
 \put(60,125){\makebox(0,0){\tt goal}}
 \put(20,70){\makebox(0,0){2}}
 \put(50,0){\line(1,2){10}}
 \put(50,0){\line(1,0){20}}
 \put(70,0){\line(-1,2){10}}
 \put(40,0){\line(1,2){20}}
 \put(40,0){\line(1,0){40}}
 \put(80,0){\line(-1,2){20}}
\end{picture}
\begin{picture}(120,150)(-250,0)
 \put(60,125){\makebox(0,0){\tt goal}}
 \put(20,70){\makebox(0,0){3}}
 \put(50,0){\line(1,2){10}}
 \put(50,0){\line(1,0){20}}
 \put(70,0){\line(-1,2){10}}
 \put(40,0){\line(1,2){20}}
 \put(40,0){\line(1,0){40}}
 \put(80,0){\line(-1,2){20}}
 \put(30,0){\line(1,2){30}}
 \put(30,0){\line(1,0){60}}
 \put(90,0){\line(-1,2){30}}
 \put(20,0){\line(1,2){40}}
 \put(20,0){\line(1,0){80}}
 \put(100,0){\line(-1,2){40}}
 \put(10,0){\line(1,2){50}}
 \put(10,0){\line(1,0){100}}
 \put(110,0){\line(-1,2){50}}
 \put(0,0){\line(1,2){60}}
 \put(0,0){\line(1,0){120}}
 \put(120,0){\line(-1,2){60}}
\end{picture}
\caption{\label{h}The head-corner parser. In order to prove that a
  string is of category {\tt goal}, the parser selects the head of the
  string (1), and proves that this element is the head-corner of the
  goal. To this end, a rule is selected of which this lexical entry is
  the head daughter.  Then the other daughters of the rule are parsed
  recursively in a bidirectional fashion: the daughters left of the
  head are parsed from right to left (starting from the head), and the
  daughters right of the head are parsed from left to right (starting
  from the head).  The result is a slightly larger head-corner (2).
  This process repeats itself until a head-corner is constructed which
  dominates the whole string (3). }
\end{figure}

\begin{figure}[h,t,b,p]\small
\begin{verbatim}
% parse(?Cat,+P0,+P)
% there is a category Cat from P0 to P
parse(Cat,P0,P) :- parse(Cat,P0,P,P0,P).

% parse(?Cat,?P0,?P,+E0,+E) 
% there is a category Cat from P0 to P within the interval E0-E
parse(Cat,P0,P,E0,E) :-
    predict(Cat,P0,P,E0,E,SmallCat,Q0,Q),
    head_corner(SmallCat,Q0,Q,Cat,P0,P,E0,E).

% head_corner(?Small,+Q0,+Q,?Cat,?P0,?P,+E0,+E)
% Small from Q0-Q is a head-corner of Cat from P0-P 
% where P0-P occurs within E0-E
head_corner(Cat,P0,P,Cat,P0,P,_,_).
head_corner(Small,Q0,Q,Cat,P0,P,E0,E) :-
    headed_rule(Small,Mother,RevLeftDs,RightDs),
    head_link(Cat,P0,P,Mother,QL,QR),
    parse_left_ds(RevLeftDs,QL,Q0,E0),  parse_right_ds(RightDs,Q,QR,E),
    head_corner(Mother,QL,QR,Cat,P0,P,E0,E).

% parse_left_ds(+RevLeftDs,-Q0,+Q,+E0) 
% there are categories LeftDs from Q0 to Q
% s.t. RevLeftDs is reverse of LeftDs, and E0=<Q0. 
parse_left_ds([],Q,Q,_).
parse_left_ds([H|T],Q0,Q,E0) :-
    parse(H,Q1,Q,E0,Q), parse_left_ds(T,Q0,Q1,E0).

% parse_right_ds(+RightDs,+Q0,-Q,+E) 
% there are categories RightDs from Q0 to Q s.t. Q =< E. 
parse_right_ds([],Q,Q,_).
parse_right_ds([H|T],Q0,Q,E) :-
    parse(H,Q0,Q1,Q0,E),  parse_right_ds(T,Q1,Q,E).

% predict(+Cat,?P0,?P,+E0,+E,-Small,-Q0,-Q)
% Small from Q0-Q (within E0-E) is a lexical category and possible
% head-corner for Cat from P0-P. 
predict(Cat,P0,P,E0,E,Small,Q0,Q) :-
    lex_head_link(Cat,P0,P,Small,Q0,Q),
    lexical_analysis(Q0,Q,Small), 
    smaller_equal(E0,Q0), 
    smaller_equal(Q,E).
\end{verbatim}
\caption{\label{speca}Definite-clause specification of the head-corner
  parser. }
\end{figure}

Head-corner parsing is a radical approach to head-driven parsing in
that it gives up the idea that parsing should proceed from left to
right. Rather, processing in a head-corner parser is
bidirectional, starting from a head outward (`island'-driven).  A
head-corner parser can be thought of as a generalisation of the
left-corner parser \cite{lc,bup,pereira-shieber}.  As in the
left-corner parser, the flow of information in a head-corner parser is
both bottom-up {\em and} top-down.

In order to explain the parser I first introduce some terminology.  I
assume that grammars are defined in the Definite Clause Grammar
formalism \cite{dcg}. Without any loss of generality I assume that no
external Prolog calls (the ones that are defined within \{ and \}) are
used, and that all lexical material is introduced in rules which have
no other right-hand-side members (these rules are called lexical
entries).  The grammar thus consists of a set of rules and a set of
lexical entries. For each rule an element of the right-hand-side is
identified as the {\em head} of that rule.  The head-relation of two
categories $h$, $m$ holds with respect to a grammar iff the grammar
contains a rule with left hand side $m$ and head daughter $h$.  The
relation `head-corner' is the reflexive and transitive closure of the
head relation.

The basic idea of the head-corner parser is illustrated in
figure~\ref{h}.  The parser selects a word (1), and
proves that the category associated with this word is the head-corner 
of the goal. To this end,
a rule is selected of which this category is the head daughter.
Then the other daughters of the rule are parsed recursively in a
bidirectional fashion: the daughters left of the head are parsed from
right to left (starting from the head), and the daughters right of the
head are parsed from left to right (starting from the head).  The
result is a slightly larger head-corner (2). This process repeats
itself until a head-corner is constructed which dominates the whole
string (3).

Note that a rule is triggered only with a fully instantiated
head-daughter.  The `generate-and-test' behaviour observed in the
previous section (examples~\ref{pra} and \ref{prb}) 
is avoided in a head-corner parser, because in the cases discussed
there, the rule
would be applied only if the {\sc vp} is found, and hence {\tt Arg} is
instantiated. For example if {\tt Arg = np(sg3,[],Subj)}, the parser
continues to search for a singular {\sc np}, and need not consider
other categories.\\

To make the definition of the parser easier, and to make sure that
rules are indexed appropriately, grammar rules are represented by the
predicate {\tt headed\_rule/4} in which the first argument is the head
of the rule, the second argument is the mother node of the rule, the
third argument is the reversed list of daughters left of the head, and
the fourth argument is the list of the daughters right of the head.
\footnote{Later we will also allow the use of rules with an empty
  right-hand-side. These will simply be represented by the predicate
  {\tt gap/1}.} This representation of a grammar will in practice be
compiled from a friendlier notation.

As an example, the DCG rule
\begin{verbatim}
x(A,E) --> a(A), b(B,A), x(C,B), d(C,D), e(D,E).
\end{verbatim}
of which the third daughter constitutes the head, is
represented now as:
\begin{verbatim}
headed_rule( x(C,B), x(A,E), [b(B,A), a(A)], [d(C,D), e(D,E)]).
\end{verbatim}

It is assumed furthermore that lexical lookup has been performed
already by another module. This module has asserted clauses for the
predicate {\tt lexical\_analysis/3} where the first two arguments
are the string positions and the third argument is the (lexical)
category.  For an input sentence `Time flies like an arrow' this
module may produce the following set of clauses:
\begin{numbered}\begin{verbatim}
lexical_analysis(0,1,verb).  lexical_analysis(0,1,noun).
lexical_analysis(0,2,noun).  lexical_analysis(1,2,noun).
lexical_analysis(1,2,verb).  lexical_analysis(2,3,prep).
lexical_analysis(2,3,verb).  lexical_analysis(3,4, det).
lexical_analysis(4,5,noun).
\end{verbatim}\end{numbered}

A simple definite-clause specification of the head-corner parser is
given in figure~\ref{speca}.  The predicate visible to the rest of the
world will be the predicate {\tt parse/3}. This predicate is defined
in terms of the {\tt parse/5} predicate. The extra arguments introduce
a pair of indices which represent the extreme positions between which
a parse should be found. This will be explained in more detail below.
A goal category can be parsed if a predicted lexical category can be
shown to be a head-corner of that goal.  The head-corner predicate
constructs (in a bottom-up fashion) larger and larger head-corners.
To parse a list of daughter categories we have to parse each daughter
category in turn.  A predicted category must be a lexical category
that lies somewhere between the extreme positions. The predicate {\tt
  smaller\_equal} is true if the first argument is a smaller or equal
integer than the second. The use of the predicates {\tt head\_link}
and {\tt lex\_head\_link} is explained below.

Note that unlike the left-corner parser, the head-corner parser may
need to consider alternative words as a possible head-corner of a
phrase, e.g. when parsing a sentence which contains several verbs.
This is a source of inefficiency if it is difficult to determine what
the appropriate lexical head for a given goal category is.  This
problem is somewhat reduced because of:
\begin{itemize}
\item the use of extremes
\item the use of top-down information
\end{itemize}

\subsection{The Use of Extremes} 
The main difference between the head-corner parser in the previous
paragraph and the left-corner parser is --- apart from the head-driven
selection of rules --- the use of two pairs of indices, to implement
the bidirectional way in which the parser proceeds through the string.

Observe that each parse-goal in the left-corner parser is
provided with a category and a left-most position. In the head-corner
parser a parse-goal is provided either with a begin or end position
(depending on whether we parse from the head to the left or to the
right) but also with the extreme positions between which the category
should be found. In general, the parse predicate is thus provided with
a category and two pairs of indices. The first pair indicates the
begin and end position of the category, the second pair indicates the
extreme positions between which the first pair should lie. 
In figure~\ref{adjp} the motivation for this technique is illustrated 
with an example. 

\begin{figure}[t,h,b]
\setlength{\unitlength}{.5pt}
\begin{picture}(300,360)(0,-20)
\put(70,20){\line(1,0){230}}
\put(70,20){\line(1,6){30}}
\put(130,20){\line(-1,6){30}}
\put(70,0){\makebox(0,0){5}}
\put(130,0){\makebox(0,0){6}}
\put(200,0){\makebox(0,0){7}}
\put(230,0){\makebox(0,0){8}}
\put(130,20){\line(1,2){85}}
\put(300,20){\line(-1,2){85}}
\put(200,20){\line(1,2){15}}
\put(230,20){\line(-1,2){15}}
\put(215,60){\makebox(0,0){{\tt n}}}
\put(185,60){\makebox(0,0){{\tt adjp}}}
\put(185,75){\line(1,2){15}}
\put(215,75){\line(-1,2){15}}
\put(100,215){\makebox(0,0){{\tt v}}}
\put(215,215){\makebox(0,0){{\tt np}}}
\put(215,230){\line(-3,1){57.5}}
\put(100,230){\line(3,1){57.5}}
\put(157.5,258){\makebox(0,0){{\tt vp}}}
\multiput(200,105)(1,5){16}{\makebox(0,0){.}}
\put(157.5,320){\makebox(0,0){{\tt s}}}
\multiput(157.5,273)(0,5){8}{\makebox(0,0){.}}
\end{picture}
\caption{\label{adjp} 
  This example illustrates how the use of two pairs of string
  positions reduces the number of possible lexical head-corners for a
  given goal.  Suppose the parser predicted (for a goal category {\tt
    s}) a category {\tt v} from position 5 to 6. In order to construct
  a complete tree {\tt s} for this head-corner, a rule is selected
  which dictates that a category {\tt np} should be parsed to the
  right, starting from position 6. To parse {\tt np}, the category
  {\tt n} from 7 to 8 is predicted. Suppose furthermore that in order
  to connect {\tt n} to {\tt np} a rule is selected which requires a
  category {\tt adjp} to the left of {\tt n}.  It will be clear that
  this category {\tt adjp} should end in position 7, but can never
  start before position 6. Hence the only candidate head-corner of
  this phrase is to be found between 6 and 7.}
\end{figure}

\subsection{Adding Top-down Filtering}
\subsubsection{Category Information}

As in the left-corner parser, a `linking' table is maintained which
represents important aspects of the head-corner relation. For some
grammars, this table simply represents the fact that the {\sc head}
features of a category and its head-corner are shared. Typically, such
a table makes it possible to predict that in order to parse a
finite sentence, the parser should start with a finite verb; to parse a
singular noun-phrase the parser should start with a singular noun, etc.

The table is defined by a number
of clauses for the predicate {\tt head\_link/2} where the first
argument is a category for which the second argument is a possible
head-corner. A sample linking table may be:
\begin{numbered}\begin{verbatim}
head_link( s,verb).  head_link(  vp, verb).
head_link( s,  vp).  head_link(  np, noun).
head_link(pp,prep).  head_link(sbar, comp).
head_link( X,   X).
\end{verbatim}\end{numbered}

\subsubsection{String Position Information}

The head-corner table also includes information about begin and end
positions, following an idea in \namecite{sikkel-diss}. For example,
if the goal is to parse a phrase with category {\em sbar} from
position 7, and within positions 7 and 12, then for some grammars it
can be concluded that the only possible lexical head-corner for this goal
should be a complementizer starting at position 7. Such information is
represented in the table as well.  This can be done by defining the
head relation as a relation between two triples, where each triple
consists of a category and two indices (representing the begin and end
position). The head relation $\langle\langle
c_m,p_m,q_m\rangle,\langle c_h,p_h,q_h\rangle\rangle$ holds iff there
is a grammar rule with mother $c_m$ and head $c_h$. Moreover, if the
list of daughters left of the head of that rule is empty, then the
begin positions are identical, i.e. $p_h = p_m$. Similarly, if the
list of daughters right of the head is empty, then $q_h = q_m$. As
before, the head-corner relation is the reflexive and transitive
closure of the head relation.

The previous example now becomes:
\begin{numbered}\label{nine}\begin{verbatim}
head_link( s,_,_, verb,_,_).  head_link(  vp,P,_,  verb,P,_).
head_link( s,_,P,   vp,_,P).  head_link(  np,_,_,  noun,_,_).
head_link(pp,P,_, prep,P,_).  head_link(sbar,P,_,  comp,P,_).
head_link( X,P,Q,    X,P,Q).
\end{verbatim}\end{numbered}

Obviously, the nature of the grammar determines whether it is useful
to represent such information. In order to be able to run a
head-corner parser in left-corner mode, this technique is
crucial. On the other hand, for grammars in which this technique does
not provide any useful top-down information no extra costs are
introduced either. 

\subsubsection{Integrating the head-corner table}
The linking table information is used to restrict 
which lexical entries are examined as candidate heads during
prediction, 
and to check whether
a rule that is selected can in fact be used in order to reach 
the current goal. To distinguish the two uses, we use the relation {\tt
  lex\_head\_link} which is a subset of the {\tt head\_link}
relation in which the head category is a possible lexical
category. An example might be the following (where we assume that the
category vp is never assigned to a lexical entry), which is a subset
of the table in \ref{nine}.
\begin{numbered}\begin{verbatim}
lex_head_link(   s,_,_, verb,_,_). lex_head_link(vp,P,_, verb,P,_).
lex_head_link(  np,_,_, noun,_,_). lex_head_link(pp,P,_, prep,P,_).  
lex_head_link(sbar,P,_, comp,P,_). lex_head_link( X,P,Q,    X,P,Q).
\end{verbatim}\end{numbered}

A few potential problems arise in connection with the use of
linking tables.  Firstly, for constraint-based grammars of the type
assumed here the number of possible non-terminals is infinite.
Therefore, we generally cannot use all information available in the
grammar but rather we should compute a `weakened' version of the
linking table.  This can be accomplished for example by replacing
all terms beyond a certain depth by anonymous variables, or by other
{\em restrictors} \cite{shieber-restriction}.

Secondly, the use of a linking table may give rise to spurious
ambiguities. Consider the case in which the category we are trying to
parse can be matched against two different items in the linking table,
but in which case the predicted head-category may turn out to be the
same.

Fortunately, the memoization technique
discussed in section~\ref{memoization} takes care of this problem.
Another possibility is to use the linking table only as a check, but
not as a source of information, by encapsulating the call within a
double negation. 
\footnote{
This approach also solves another potential problem: the linking table
may give rise to (undesired) cyclic terms due to the absence of the
occur check. The double negation takes care of this potential problem
too. }

The solution implemented in the head-corner parser is to use, for each
pair of functors of categories, the generalization of the head-corner
relation. Such functors typically are major and minor syntactic
category labels such as {\sc np, vp, s, sbar, verb \dots}. As a result
there will always be at most one matching clause in the linking table
for a given goal category and a given head category (thus there is no
risk of obtaining spurious ambiguities). Moreover, this approach
allows a very efficient implementation technique which is described
in the following paragraph.

\subsubsection{Indexing of the head-corner table}

In the implementation of the head-corner parser we use an efficient
implementation of the head-corner relation by exploiting Prolog's
first argument indexing. This technique ensures that the lookup of the
head-corner table can be done in (essentially) constant time. The
implementation consists of two step. In the first step the head-corner
table is weakened such that for a given goal category and a given head
category at most a single matching clause exists. In the second step
this table is encoded in such a way that first argument indexing
ensures that table lookup is efficient.

As a first step we modify the head-corner relation to make sure that
for all pairs of functors of categories, there will be at most one
matching clause in the head-corner table. This is illustrated with an
example. Suppose a hypothetical head-corner table contains the
following two clauses relating categories with functor {\tt x/4} and
{\tt y/4}:
\begin{verbatim}
head_link(x(A,B,_,_),_,_,y(A,B,_,_),_,_).
head_link(x(_,B,C,_),_,_,y(_,B,C,_),_,_).
\end{verbatim}
In this case, the modified head-corner relation table will consist of
a single clause relating {\tt x/4} and {\tt y/4} by taking the
generalization (or `anti-unification') of the two clauses:
\begin{verbatim}
head_link(x(_,B,_,_),_,_,y(_,B,_,_),_,_).
\end{verbatim}
As a result, for a given goal and head category, table lookup is
deterministic. 

In the second and final step of the modification we re-arrange the
information in the table such that for each possible goal category 
functor {\tt g/n} there will be a clause: 
\begin{verbatim}
head_link(g(A1..An),Pg,Qg,Head,Ph,Qh) :-
    head_link_G_N(Head,Ph,Qh,g(A1..An),Pg,Qg).
\end{verbatim}
Moreover, all the relations {\tt head\_link\_G\_N} now contain the
relevant information from the head-corner table. Thus, for
clauses of the form:
\begin{verbatim}
head_link(x(_,B,_,_),_,_,y(_,B,_,_),_,_).
\end{verbatim}
we now have:
\begin{verbatim}
head_link_x_4(y(_,B,_,_),_,_,x(_,B,_,_),_,_).
\end{verbatim}
First argument indexing now ensures that table lookup is efficient.

The same technique is applied for the {\tt lex\_head\_link} relation. 
This technique significantly improved the practical time efficiency of
the parser (especially if the resulting code is compiled).

\subsection{Dealing with Epsilon Rules}

In the preceding paragraphs we have said nothing about empty
productions (epsilon rules). A possible approach is to compile the
grammar into an equivalent grammar in which no such epsilon rules are
defined. It is also possible to deal with epsilon rules in the
head-corner parser directly. For example, we could assert
empty productions as possible `lexical analyses'. In
such an approach the result of lexical analysis may contain clauses
such as the following, in case there is a rule \verb+ np/np --> []+. 
\begin{numbered}\begin{verbatim}
lexical_analysis(0,0,np/np). lexical_analysis(1,1,np/np).
lexical_analysis(2,2,np/np). lexical_analysis(3,3,np/np).
lexical_analysis(4,4,np/np).
\end{verbatim}\end{numbered}

There are two objections to this approach. The first objection may
be that this is a task that can hardly be expected from a lexical
lookup procedure. The second, more important, objection is that empty
categories are hypothesized essentially everywhere. 

In the general version of the head-corner parser gaps are inserted by
a special clause for the {\tt predict/8} predicate (\ref{prgap}) where
shared variables are used to indicate the corresponding string positions.
The {\tt gap\_head\_link} relation is a subset of the {\tt head\_link}
relation in which the head category is a possible gap.
\begin{numbered}\label{prgap}\begin{verbatim}
predict(Cat,P0,P,_E0,_E,Small,Q,Q) :-
    gap_head_link(Cat,P0,P,Small,Q,Q),
    gap(Small).
\end{verbatim}\end{numbered}
In order for this approach to work other predicates must expect string
positions which are not instantiated.  For example, Prolog's built-in
comparison operator cannot be used, since that operator requires that
its arguments are ground.  The definition of the {\tt smaller\_equal}
predicate therefore reflects the possibility that a string position is
a variable (in which case calls to this predicate should succeed).

For some grammars it turns out that a simplification is possible. If
it is never possible that a gap can be used as the head of a
rule, then we can omit this new clause for the {\tt predict}
predicate, and instead use a new clause for the {\tt parse/5}
predicate, as follows:
\begin{numbered}\label{gap}\begin{verbatim}
parse(Small,Q,Q,_E0,_E) :-
    gap(Small).
\end{verbatim}\end{numbered}
This will typically be much more efficient because in this case gaps
are hypothesized in a purely top-down manner.\\

It should be noted that the general version of the head-corner parser is
not guaranteed to terminate, even if the grammar defines only a finite
number of derivations for all input sentences. Thus, even though the
head-corner parser proceeds in a bottom-up direction, it can run into
left-recursion problems (just like the left-corner parser can).  This
is because it may be possible that an empty category is predicted as
the head, after which trying to construct a larger projection of this
head gives rise to a parse goal for which a similar empty category is
a possible candidate head \dots. This problem is sometimes called
`hidden left-recursion' in the context of left-corner parsers. 

This problem can be solved in some cases by a good (but relatively
expensive) implementation of the memoization technique, e.g. along the
lines of \namecite{ds-warren} or \namecite{johnson-doerre}.  The
simplified (and more efficient) memoization technique that I use (cf.\ 
section~\ref{memoization}) however does not solve this problem.

A quite different solution, which is often applied for the same
problem if a left-corner parser is used, is to compile the grammar
into an equivalent grammar without gaps. For left-corner parsers this
can be achieved by partially evaluating all rules which can take
gap(s) as their left-most daughter(s). Therefore, the parser only
needs to consider gaps in non-leftmost position by a clause similar to
the clause in~\ref{gap}. Obviously, the same compilation technique can
be applied for the head-corner parser too. However, there is a
problem: it will be unclear what the heads of the newly created rules
will be.  Moreover and more importantly, the head-corner relation will
typically become much less predictive. For example, if there is a rule
\verb+vp --> np verb+ where the {\tt verb} can be realized as a gap,
then after compilation a rule of the form {\tt vp --> np} will exist.
Therefore, a {\tt np} will be a possible head-corner of {\tt vp}. The
effect will be that head-corners are difficult to predict, and hence
efficiency decreases. 

Experience suggests that grammars exhibiting `hidden head-recursion'
can often be avoided.  \footnote{ For example, in the Alvey NL Tools
  grammar in only 3 (out of more than 700) rules the head of the rule
  could be gapped.  These rules are of the form \verb+x --> not x+.
  Arguably, in such rules the second daughter should not be gapped.
  In the MiMo2 grammar of English, no heads can be gapped. Finally, in
  the Dutch OVIS grammar (in which verb-second is implemented by
  gap-threading) no hidden head-recursion occurs, as long as the
  head-corner table includes information about the feature {\tt
    vslash}, which encodes whether or not a v-gap is expected. }

\section{Selective Memoization and Goal-weakening}
\label{memoization}

\subsection{Selective Memoization}
The basic idea behind memoization is simple: do not compute things
twice. In Prolog we can keep track of each goal that
has already been searched and keep a list of the corresponding
solution(s). If the same goal needs to be solved later, then we can
skip the computation and simply do a table lookup.  The cost of
maintaining a table and doing the table lookup is rather expensive
itself.  Therefore, we should modify the slogan `do not compute things
twice' into: `do not compute {\em expensive} things twice'.

In the head-corner parser it turns out that the {\tt parse/5} predicate
is a very good candidate for memoization. The other predicates are
not. This implies that each {\em maximal projection} is computed only
once; partial projections of a head can be constructed during a parse
any number of times, as can sequences of categories (considered as
sisters to a head). Active chart parsers `memo' everything (including
sequences of categories); inactive
chart parsers only memo categories, but not sequences of
categories. In our proposal, we memo only those categories that are
`maximal projections', i.e. projections of a head which unify with the
top category (start symbol) or with a non-head daughter of a rule. \\

The implementation of memoization uses Prolog's
internal database to store the tables. The advantage of this technique
is that we use Prolog's first argument indexing for such
tables. Moreover, during the consultation of the table we need not
worry about modifications to it (in contrast to an approach
in which the table would be maintained as the value of a Prolog
variable).  On the other hand, the use of the internal database brings
about a certain overhead. Therefore, it may be worthwhile to
experiment with a meta-interpreter along the lines of the XOLDT system
\cite{ds-warren} in which the table is maintained dynamically. \\ 

Memoization is implemented by two different tables. The first table
encodes which goals have already been searched. Items in the first table
are called {\em goal items}. The second table
represents all solved (i.e. instantiated) goals. Items in this second
table are called {\em result items}. One may be tempted
to use only the second table. But in that case we would not be able to
tell the difference between a goal which has already been searched,
but did not result in a solution (`fail-goal') and a goal which has
not been searched at all. If we have two tables then we can also
immediately stop working on branches in the search space for which it
has already been shown that there is no solution. The distinction
between these two kinds of item is inherited from BUP \cite{bup}. The
memoized version of the parse predicate can be defined as
in~(\ref{memoparse}).

\begin{numbered}\label{memoparse}\begin{verbatim}
parse(Cat,P0,P,E0,E) :-
    (  in_table1(Cat,P0,P,E0,E)                 % done before?
    -> true                                     % then don't search
    ;  ( predict(Cat,P0,P,E0,E,SmCat,Q0,Q),     % otherwise find all
         head_corner(SmCat,Q0,Q,Cat,P0,P,E0,E), % results and assert
         assert_table2(Cat,P0,P),               % these
         fail
       ; assert_table1(Cat,P0,P,E0,E)           % goal is now done
    )  ),
    in_table2(Cat,P0,P,E0,E).                   % pick a solution
\end{verbatim}\end{numbered}

The first table is represented by the predicate {\tt 'GOAL\_ITEM'}.
This predicate simply consists of a number of unit-clauses indicating
all goals that have been searched completely. Thus, before we try to
attempt to solve {\tt Goal} we first check whether a goal item for
that goal already exists. Given the fact that {\tt Goal} may contain
variables we should be a bit careful here. Unification is clearly not
appropriate here since that may result in a situation in which a more
general goal is not searched because a more specific variant of that
goal had been solved. We want exactly the opposite: if a more general
version of {\tt Goal} is included in the goal table,
then we can continue to look for a solution in the result table.
It is useful to consider the fact that if we had previously
solved e.g. the goal {\tt parse(s,3,X,3,12)}, then if we later
encounter the goal {\tt parse(s,3,Y,3,10)}, we can also use the second
table immediately: the way in which the extreme positions are used
ensures that the former is more general than the latter. The
predicates for the maintenance of the goal table are defined in~(\ref{mg}). 
\begin{numbered}\label{mg}\begin{verbatim}
in_table1(Cat,P0,P,E0,E) :-
    'GOAL_ITEM'(Cat_d,P0_d,P_d,E0_d,E_d),       % goal exists which is
    subsumes_chk((Cat_d,P0_d,P_d),(Cat,P0,P)),  % more general and within
    smaller_equal(E0_d,E0),                     % a larger interval
    smaller_equal(E,E_d).

assert_table1(Cat,P0,P,E0,E) :- assertz('GOAL_ITEM'(Cat,P0,P,E0,E)).
\end{verbatim}\end{numbered}

The second table is represented by the predicate {\tt 'RESULT\_ITEM'}.
It is defined by unit-clauses which each represent an instantiated
goal (i.e. a solution). Each time a result is found, it is checked
whether that result is already available in the table. If so, the
newer result is ignored. If no (more general version of) the result
existed, then the result is added to the table. Moreover, more
specific results which may have been put on the table previously are
marked. These results need not be used anymore. \footnote{Note that
  such items are not removed, because in that case the item reference
  becomes available for later items, which is unsound. } This is not
strictly necessary but is often useful because it decreases the size
of the tables; in this approach tables are redundancy-free and hence
minimal.  Moreover, such more specific results cannot be used anymore
and no work will be done based on those results.  Note that
RESULT\_ITEMs do not keep track of the extreme positions. This implies
that in order to see whether a RESULT\_ITEM is applicable we check
whether the interval covered by the RESULT\_ITEM lies within the
extreme positions of the current goal.  The predicates dealing with
the result table are defined in~(\ref{rg}).

\begin{numbered}\label{rg}\begin{verbatim}
in_table2(Cat,P0,P,E0,E) :-                     
    clause('RESULT_ITEM'(Cat,P0,P),Ref),        % result exists, not
    \+ 'REPLACED_ITEM'(Ref,_),                  % replaced by general result
    smaller_equal(E0,P0),  smaller_equal(P,E).  % within desired interval

assert_table2(Cat,P0,P):-
    (  'RESULT_ITEM'(Cat_d,P0_d,P_d),           % if result exists
       subsumes_chk((Cat_d,P0_d,P_d),(Cat,P0,P) % which is more general
    -> true                                     % then ok
    ;  assertz('RESULT_ITEM'(Cat,P0,P),Ref),    % otherwise assert it, and
       mark_item('RESULT_ITEM'(Cat,P0,P),Ref)   % mark more specific items
    ).

mark_item(Cat,NewRef) :-
     ( clause(Specific,_,Ref),                  % item exists
       \+ Ref=NewRef,                           % not the one just added
       subsumes_chk(Cat,Specific),              % and it's more specific
       assertz('REPLACED_ITEM'(Ref,NewRef)),    % then mark it
       fail                                     % do this for all such items
     ; true 
     ).
\end{verbatim}\end{numbered}

The implementation uses a faster implementation of memoizating in
which both goal items and result items are indexed by the functor of
the category and the string positions.

In the head-corner parser, parse goals are memoized.  Note that
nothing would prevent us from memoing other predicates as well.  Experience
suggests that the cost of maintaining tables for e.g. the {\tt
  head\_corner} relation is (much) higher than the associated profit.
The use of memoization for only the {\tt parse/5} goals implies that
the memory requirements of the head-corner parser in terms of the
number of items that is being recorded is much smaller than in
ordinary chart parsers. Not only do we refrain from asserting
so-called {\em active} items, but we also refrain from asserting {\em
  inactive} items for non-maximal projections of heads. In practice
the difference in space requirements can be enormous.  This difference
is a significant reason for the practical efficiency of
the head-corner parser.  

\subsection{The Occur Check}

It turns out that the use of tables defined in the previous subsection
can lead to a problem with cyclic unifications. If we assume that
Prolog's unification includes the occur check then no problem would
arise. But since most versions of Prolog do not implement the
occur check it is worthwhile investigating this potential problem. 

The problem arises because cyclic solutions can be constructed that
would not have been constructed by ordinary
SLD-resolution. Furthermore, these cyclic structures lead to practical
problems because items containing such a cyclic structure may have to
be put in the table. In SICStus Prolog this results in a crash. 

An example may clarify the problem. Suppose we have a very simple
program containing the following unit clause:
\begin{verbatim}
x(A,B).
\end{verbatim}
Furthermore suppose that in the course of the computation a goal of
the form 
\begin{verbatim}
?- x(f(X),X)
\end{verbatim}
is attempted. This clearly succeeds. Furthermore an item of that form
is added to the table. Later on it may be the case that a goal of
the form 
\begin{verbatim}
?- x(Y,Y)
\end{verbatim}
is attempted. Clearly this is not a more specific goal than we solved
before, so we need to solve this goal afresh. This succeeds too. Now
we can continue by picking up a solution from the second
table. However, if we pick the first solution then a cyclic term
results. 

A possible approach to deal with this situation is to index the items
of the second table with the item of the first table from which the
solution was obtained. In other words: if you want to select a
solution from the second table, it must not only be the case that the
solution matches your goal, but also that 
the corresponding goal of the solution is more general than your
current goal. This strategy works, but turns out to be considerably
slower than the original version given above. The reason seems to be
that the size of the second table is increased quite drastically,
because solutions may now be added to the table more than once (for
all goals that could give rise to that solution).\\

It turns out that an improvement of the head-corner parser using a
{\em goal weakening} technique often eliminates this occur check
problem. Goal weakening is discussed in the following subsection.

\subsection{Goal Weakening}

The insight behind `goal weakening' (or {\em abstraction} 
\cite{johnson-doerre}) in the
context of memoization is that we may combine a number of slightly
different goals into a single more general goal. Very often it is much
cheaper to solve this single (but more general) goal, than to solve
each of the specific goals in turn. Moreover, the goal table will be
smaller (both in terms of number of items, and the size of individual
items), which can have a very good effect on the amount of memory and
CPU-time required for the administration of the table.  Clearly, one
must be careful not to remove essential information from the goal (in
the worst case this may even lead to non-termination of otherwise
well-behaved programs).

Depending on the properties of a particular grammar, it may
for example be worthwhile to {\em restrict} a given category to its
syntactic features before we attempt to solve the parse goal of that
category. Shieber's restriction operator \cite{shieber-restriction} can be
used here. Thus we essentially throw some information away before an
attempt is made to solve a (memoized) goal.  For example, the category
\begin{verbatim}
x(A,B,f(A,B),g(A,h(B,i(C))))
\end{verbatim}
may be weakened into:
\begin{verbatim}
x(A,B,f(_,_),g(_,_))
\end{verbatim}
If we assume that the predicate {\tt weaken/2} relates a term $t$
to a weakened version $t_w$, such that $t_w$ subsumes $t$, then 
(\ref{mww}) is the improved version of the parse predicate:
\begin{numbered}\label{mww}\begin{verbatim}
parse_with_weakening(Cat,P0,P,E0,E) :-
    weaken(Cat,WeakenedCat),
    parse(WeakenedCat,P0,P,E0,E),
    Cat=WeakenedCat.
\end{verbatim}\end{numbered}

Note that goal weakening is sound. An answer $a$ to a weakened goal $g$ is
only considered if $a$ and $g$ unify. Also note
that goal-weakening is complete in the sense that for an answer $a$ to
a goal $g$ there will always be an answer $a'$ to the weakening of $g$
such that $a'$ subsumes $a$. 

For practical implementations the use of goal weakening can be
extremely important. It is my experience that a well-chosen goal
weakening operator may reduce parsing times by an order of
magnitude. \\

The goal weakening technique can also be used to eliminate typical
instances of the problems concerning the occur check (discussed in
the previous subsection). Coming back to the example in the previous
subsection, if our first goal 
\begin{verbatim}
x(f(X),X)
\end{verbatim}
were weakened into
\begin{verbatim}
x(f(_),_)
\end{verbatim}
then the problem would not occur. If we want to guarantee that no
cyclic structures can be formed then we would need to define
goal-weakening in such a way that no variable sharing occurs in
the weakened goal. 

An important question is how to come up with a good goal
weakening operator. For the experiments discussed in the final section
all goal weakening operators were chosen by hand, based on small
experiments and inspection of the goal table and item table. Even if
goal-weakening is reminiscent of Shieber's restriction operator
\cite{shieber-restriction}, the rules of the game are quite different:
in the former case as much information as possible is removed 
without risking non-termination of the parser.
In the latter case
information is removed {\em until} the resulting parser terminates.
For the current version of the grammar of OVIS, it turned out that
weakening the goal category in such a way that all information below a
depth of 6 is replaced by fresh variables eliminated the problem caused
by the absence of the occur check; moreover this goal weakening
operator reduced parsing times substantially. In the latest version we
use different goal weakening operators for each different functor.

An interesting special case of goal-weakening is constituted by a
goal-weakening operator which ignores all feature constraints, and
hence only leaves the functor for each goal category. In this case the
administration of the goal table can be simplified considerably (the
table consists of ground facts, hence no subsumption checks are
required). This technique is used in the MiMo2 grammar and the Alvey
NL Tools grammar (both discussed in section~\ref{compar}).

\section{Compact Representation of Parse Trees}

Often a distinction is made between {\em recognition} and
{\em parsing}. Recognition checks whether a given sentence can be
generated by a grammar. Usually recognizers can be  adapted to
be able to recover the possible parse trees of that sentence (if
any).

In the context of Definite-clause Grammar this
distinction is often blurred because it is possible to build up the
parse tree as part of the complex non-terminal symbols. Thus the
parse tree of a sentence may be constructed as a side-effect of the
recognition phase. If we are interested in logical forms
rather than in parse trees a similar trick may be used. The result
of this however is that already during recognition ambiguities will
result in a (possibly exponential) increase of processing time. 

For this reason we will assume that parse trees are {\em not} built by
the grammar, but rather are the responsibility of the parser. This
allows the use of efficient {\em packing} techniques. The result of
the parser will be a {\em parse forest}: a compact representation of 
all possible parse trees rather than an enumeration of all parse
trees. \\

The structure of the `parse-forest' in the head-corner parser is
rather unusual, and therefore we will take some time to explain it.
Because the head-corner parser uses selective memoization,
conventional approaches to construct parse forests \cite{billot-lang}
are not applicable.  The head-corner parser maintains a table of
partial derivation-trees which each represent a successful path from a
lexical head (or gap) up to a goal category. The table consisting of
such partial parse trees is called the history table; its items are
history-items.

More specifically, each history-item is a triple consisting of a
result-item reference, a rule name and a list of triples. The rule
name is always the name of a rule without daughters (i.e. a lexical
entry or a gap): the (lexical) head. Each triple in the list of
triples represents a local tree. It consists of the rule name, and two
lists of result-item references (representing the list of daughters
left of the head in reverse, and the list of daughters right of the
head).  An example will clarify this. Suppose we have a history-item:
\begin{numbered}\begin{verbatim}
'HISTORY_ITEM'(112,give22,
    [rule(vp_v,[],[]),      rule(vp_vp_np_pp,[],[121,125]),
     rule(s_np_vp,[87],[]), rule(s_adv_s,[46],[])]).
\end{verbatim}\end{numbered}
\begin{figure}[h,t,b]
\centerline{
\leavevmode
\unitlength1pt
\picture(115.50,158.57)
\catcode`\@=11
 \put(6.42,138.00){\hbox{112:s-adv-s}}
 \put(26.37,131.79){\hbox{}}
 \put(23.1536,134.7850){\lineseg(-1,-1){16.8051}}
 \put(0.00,108.00){\hbox{46}}
 \put(5.00,101.79){\hbox{}}
 \put(29.0477,134.7850){\lineseg(5,-6){15.8917}}
 \put(30.54,108.00){\hbox{s-np-vp}}
 \put(47.74,99.18){\hbox{}}
 \put(41.9173,102.1800){\lineseg(-1,-1){14.3502}}
 \put(19.42,78.00){\hbox{87}}
 \put(24.42,71.79){\hbox{}}
 \put(53.5572,102.1800){\lineseg(1,-1){16.4651}}
 \put(49.97,78.00){\hbox{vp-vp-np-pp}}
 \put(71.05,69.18){\hbox{}}
 \put(62.3208,72.1800){\lineseg(-3,-2){24.6976}}
 \put(23.78,48.00){\hbox{vp-v}}
 \put(34.10,39.18){\hbox{}}
 \put(34.0986,42.1800){\lineseg(0,-1){14.3502}}
 \put(19.64,18.00){\hbox{give22}}
 \put(34.10,9.18){\hbox{}}
 \put(71.0507,72.1800){\lineseg(0,-1){14.2001}}
 \put(64.96,48.00){\hbox{121}}
 \put(72.46,42.00){\hbox{}}
 \put(80.7506,72.1800){\lineseg(5,-3){23.6669}}
 \put(100.50,48.00){\hbox{125}}
 \put(108.00,41.79){\hbox{}}
 \put(0.00,135.00){\hbox{}}
\endpicture
}
\caption{\label{pt}Example of a partial derivation-tree projected by a
  history-item.}
\end{figure}
This item indicates that there is a possible derivation of the
category defined in result-item 112 of the
form illustrated in figure~\ref{pt}. In this figure the labels of the
interior nodes are rule-names, and the labels of the leaves are
references to result-items. The head-corner leaf is special: it is a
reference to either a lexical entry or an epsilon rule. The root node
is special too: it has both an associated rule name and a reference to
a result item. The latter indicates how this partial derivation tree
combines with other partial trees. 

The history table is a lexicalized tree substitution grammar, in which
all nodes (except substitution nodes) are associated with a rule
identifier (of the original grammar). 
This grammar derives exactly all derivation trees of the input.
\footnote{The tree substitution grammar is lexicalized in the sense
  that each of the trees has an associated anchor, which is a pointer
  to either a lexical entry or a gap. } As an example, consider the
grammar which is used by \namecite{tomita} and \namecite{billot-lang},
given here in~(\ref{grm}) and ~(\ref{lxx}).
\begin{numbered}\label{grm}\begin{verbatim}
(1) s --> np, vp.       (2) s --> s, pp.    (3) np --> n.       
(4) np --> det, n.      (5) np --> np, pp.  (6) pp --> prep, np.
(7) vp --> v, np.
\end{verbatim}\end{numbered}
\begin{numbered}\label{lxx}\begin{verbatim}
n    --> ['I'].         n   --> [man].      v --> [see].    
prep --> [at].          det --> [a].        n --> [home].
\end{verbatim}\end{numbered}
The sentence `I see a man at home' has two derivations, according to
this grammar. The lexicalized tree substitution grammar in
figure~\ref{tr}, which is constructed by the head-corner parser,
derives exactly these two derivations.

\begin{figure}[h,t,b]
\vspace{-5em}

\begin{center}
nt5:I ~~~     nt0:a ~~~ 
\message{including picture made by 'Treemaker' V1.232 (c)1992 University of Paderborn}
\leavevmode
\unitlength1pt
\picture(48.54,61.89)(0,44)
\catcode`\@=11
 \put(12.12,43.00){\hbox{nt1:4}}
 \put(23.09,37.00){\hbox{}}
 \put(20.5903,40.0000){\lineseg(-5,-6){10.4630}}
 \put(0.00,18.00){\hbox{nt0}}
 \put(7.08,12.00){\hbox{}}
 \put(25.3403,40.0000){\lineseg(3,-4){11.0208}}
 \put(29.65,18.00){\hbox{man}}
 \put(39.10,12.00){\hbox{}}
 \put(0.00,40.00){\hbox{}}
\endpicture
~~~
\message{including picture made by 'Treemaker' V1.232 (c)1992 University of Paderborn}
\leavevmode
\unitlength1pt
\picture(23.33,61.89)(0,44)
\catcode`\@=11
 \put(0.69,43.00){\hbox{nt2:3}}
 \put(11.67,37.00){\hbox{}}
 \put(11.6667,40.0000){\lineseg(0,-1){12.0556}}
 \put(0.00,18.00){\hbox{home}}
 \put(11.67,12.00){\hbox{}}
 \put(39.10,15.00){\hbox{}}
\endpicture
~~~
\message{including picture made by 'Treemaker' V1.232 (c)1992 University of Paderborn}
\leavevmode
\unitlength1pt
\picture(38.54,61.89)(0,44)
\catcode`\@=11
 \put(6.98,43.00){\hbox{nt3:6}}
 \put(17.95,37.00){\hbox{}}
 \put(15.7014,40.0000){\lineseg(-3,-4){9.6369}}
 \put(0.00,18.00){\hbox{at}}
 \put(4.44,12.00){\hbox{}}
 \put(20.2014,40.0000){\lineseg(3,-4){9.4167}}
 \put(24.38,18.00){\hbox{nt2}}
 \put(31.46,12.00){\hbox{}}
 \put(0.00,40.00){\hbox{}}
\endpicture

\message{including picture made by 'Treemaker' V1.232 (c)1992 University of Paderborn}
\leavevmode
\unitlength1pt
\picture(55.24,86.89)
\catcode`\@=11
 \put(24.65,68.00){\hbox{nt4:5}}
 \put(35.63,62.00){\hbox{}}
 \put(33.6251,65.0000){\lineseg(-2,-3){8.3704}}
 \put(20.59,43.00){\hbox{4}}
 \put(23.09,37.00){\hbox{}}
 \put(20.5903,40.0000){\lineseg(-5,-6){10.4630}}
 \put(0.00,18.00){\hbox{nt0}}
 \put(7.08,12.00){\hbox{}}
 \put(25.3403,40.0000){\lineseg(3,-4){11.0208}}
 \put(29.65,18.00){\hbox{man}}
 \put(39.10,12.00){\hbox{}}
 \put(37.6251,65.0000){\lineseg(2,-3){8.3704}}
 \put(41.08,43.00){\hbox{nt3}}
 \put(48.16,37.00){\hbox{}}
 \put(0.00,65.00){\hbox{}}
\endpicture
~~~
\message{including picture made by 'Treemaker' V1.232 (c)1992 University of Paderborn}
\leavevmode
\unitlength1pt
\picture(53.73,86.89)
\catcode`\@=11
 \put(8.65,68.00){\hbox{nt6:1}}
 \put(19.62,62.00){\hbox{}}
 \put(17.6181,65.0000){\lineseg(-2,-3){8.3704}}
 \put(0.00,43.00){\hbox{nt5}}
 \put(7.08,37.00){\hbox{}}
 \put(21.6181,65.0000){\lineseg(2,-3){8.3704}}
 \put(29.65,43.00){\hbox{7}}
 \put(32.15,37.00){\hbox{}}
 \put(30.1528,40.0000){\lineseg(-2,-3){9.7963}}
 \put(11.24,18.00){\hbox{see}}
 \put(17.66,12.00){\hbox{}}
 \put(34.5528,40.0000){\lineseg(4,-5){10.0444}}
 \put(39.56,18.00){\hbox{nt4}}
 \put(46.65,12.00){\hbox{}}
 \put(0.00,65.00){\hbox{}}
\endpicture
~~~
\message{including picture made by 'Treemaker' V1.232 (c)1992 University of Paderborn}
\leavevmode
\unitlength1pt
\picture(53.73,111.89)(0,25)
\catcode`\@=11
 \put(21.18,93.00){\hbox{nt6:2}}
 \put(32.15,87.00){\hbox{}}
 \put(30.1529,90.0000){\lineseg(-2,-3){8.3704}}
 \put(17.12,68.00){\hbox{1}}
 \put(19.62,62.00){\hbox{}}
 \put(17.6181,65.0000){\lineseg(-2,-3){8.3704}}
 \put(0.00,43.00){\hbox{nt5}}
 \put(7.08,37.00){\hbox{}}
 \put(21.6181,65.0000){\lineseg(2,-3){8.3704}}
 \put(29.65,43.00){\hbox{7}}
 \put(32.15,37.00){\hbox{}}
 \put(30.1528,40.0000){\lineseg(-2,-3){9.7963}}
 \put(11.24,18.00){\hbox{see}}
 \put(17.66,12.00){\hbox{}}
 \put(34.5528,40.0000){\lineseg(4,-5){10.0444}}
 \put(39.56,18.00){\hbox{nt1}}
 \put(46.65,12.00){\hbox{}}
 \put(34.1529,90.0000){\lineseg(2,-3){8.3704}}
 \put(37.60,68.00){\hbox{nt3}}
 \put(44.69,62.00){\hbox{}}
 \put(0.00,90.00){\hbox{}}
\endpicture
\end{center}

\caption{\label{tr}Tree substitution grammar which derives each of the
  two derivation trees of the sentence `I see a man at home', for the
  grammar of Billot and Lang (1994). The start symbol of this grammar
  is {\tt nt6}. Note that all nodes, except for substitution nodes, are
  associated with a rule (or lexical entry) of the original grammar.
  Root nodes have a non-terminal symbol before the colon,
  and the corresponding rule identifier after the colon. 
  The set of derived trees for this tree substitution grammar equals
  the set of derivation trees of the parse (ignoring the non-terminal
  symbols of the tree substution grammar). }
\end{figure}

Note that the item references are used in the same manner as the
computer generated names of non-terminals in the approach of
\namecite{billot-lang}.  Because we use chunks of parse trees
less packing is possible than in their approach. Correspondingly, the
theoretical worst-case space requirements are worse too. In practice,
however, this doesn't seem to be problematic at all: in our
experiments the size of the history table is always much smaller than
the size of the other tables (this is expected because the latter
tables have to record complex category information). \\ 

Let us now look at how the parser of the previous section can be adapted
to be able to assert history-items. Firstly we add an (output-)
argument to the {\tt parse} predicate. This sixth argument is the
reference to the result-item that was actually used. The predicates to
parse a list of daughters are augmented with a list of such
references. This enables the construction of a term for each local
tree in the head\_corner predicate consisting of the name of the rule
that was applied and the list of references of the result-items used
for the left and right daughters of that rule. Such a local tree
representation is an element of a list that is maintained for each
lexical head upward to its goal. Such a list thus represents in a
bottom-up fashion all rules and result-items that were used to show
that that lexical entry indeed was a head-corner of the goal. If a
parse goal has been solved then this list containing the history
information is asserted in a new kind of table: the {\tt
  'HISTORY\_ITEM'/3} table.  \footnote{A complication is needed for
  those cases where items are removed later because a more general
  item has been found. }\\ 

We already argued above that parse trees should not be explicitly
defined in the grammar. Logical forms often implicitly represent
the derivational history of a category.  Therefore, the
common use of logical forms as part of the categories will imply that
you will hardly ever find two different analyses for a single
category, because two different analyses will also have two
different logical forms.  Therefore, no packing is possible and the
recognizer will behave as if it is enumerating all parse trees.
The solution to this problem is to delay the evaluation of semantic
constraints. During the first phase all constraints referring to
logical forms are ignored. Only if a parse tree is recovered from the
parse-forest we add the logical form constraints. This is similar to
the approach worked out in CLE \cite{cle-book}.

This approach may lead to a situation in which the second phase
actually filters out some otherwise possible derivations, in case the
construction of logical forms is not {\em compositional} in the
appropriate sense. In such cases the first phase may be said to be
unsound in that it allows ungrammatical derivations. The first phase
combined with the second phase is of course still sound. Furthermore,
if this situation arose very often, then the first phase would
tend to be useless, and all work would have to be done during the
recovery phase.  The present architecture of the head-corner parser
embodies the assumption that such cases are rare, and that the
construction of logical forms is (grosso modo) compositional.

The distinction between semantic and syntactic information is compiled
into the grammar rules on the basis of a user declaration. We simply
assume that in the first phase the parser only refers to syntactic
information, whereas in the second phase both syntactic and semantic
information is taken into account. 

If we assume that the grammar constructs logical forms, then it is not
clear that we are interested in parse trees at all. A simplified
version of the recover predicate may be defined in which we only
recover the semantic information of the root category, but in which we
don't build parse trees. The simplified version may be regarded as
the run-time version, whereas parse trees will still be very useful
for grammar development.  

\section{Parsing Word-graphs with Probabilities}
\label{wordgraphs}
The head-corner parser is one of the parsers developed within the NWO
Priority Programme on Language and Speech Technology. In this program a
spoken dialog system is developed for public transportation
information \cite{ovisplan}.

In this system the input for the parser is not a simple list of words,
as we have assumed up to now, but rather a word-graph: a directed,
acyclic graph where the states are points in time, and the edges are
labelled with word hypotheses and their corresponding acoustic score.
Thus, such word-graphs are acyclic weighted finite-state automata. 

In \namecite{lang-atr} a framework for processing ill-formed input is
described in which certain common errors are modelled as (weighted)
finite-state transducers. The composition of an input sentence with
these transducers produces a (weighted) finite state automaton which
is then input for the parser. In such an approach the need to
generalize from input strings to input finite-state automata 
is also clear.

The generalization from strings to weighted acyclic finite-state automata
introduces essentially two complications. Firstly, we cannot use
string indices anymore. Secondly we need to keep track of the
acoustic scores of the words used in a certain derivation.

\subsection{From string positions to state names}
Parsing on the basis of a finite-state automaton can be seen as the
computation of the intersection of that automaton with the grammar. 
If the definite-clause grammar is off-line
parsable, and if the finite-state automaton is acyclic, then this
computation can be guaranteed to terminate
\cite{acl95}. 
This is obvious because an acyclic finite-state
automaton defines a finite number of strings. 
More importantly, existing techniques for parsing based on
strings can be generalized easily by using the names of states in the
automaton instead of the usual string indices. 

In the head-corner parser, this leads to an alternative to
the predicate {\tt smaller\_equal/2}.  Rather than a simple integer
comparison, we now need to check that a derivation from {\tt P0} to
{\tt P} can be extended to a derivation from {\tt E0} to {\tt E} by
checking that there are paths in the word-graph from {\tt E0} to {\tt
  P0} and from {\tt P} to {\tt E}. 

The predicate {\tt connection/2} is true if there is a path in the
word-graph from the first argument to the second argument.  It is
assumed that state names are integers; to rule out cyclic word-graphs
we also require that for all transitions from {\tt P0} to {\tt P} it
is the case that {\tt P0} $<$ {\tt P}.  Transitions in the word-graph
are represented by clauses of the form {\tt
  wordgraph:trans(P0,Sym,P,Score)} which indicate that there is a
transition from state {\tt P0} to {\tt P} with symbol {\tt Sym} and
acoustic score {\tt Score}. The connection predicate can be specified
simply as the reflexive and transitive closure of the transition
relation between states:
\begin{numbered}\begin{verbatim}
connection(A,A).
connection(A0,A) :-
        wordgraph:trans(A0,_,A1,_),
        connection(A1,A).
\end{verbatim}\end{numbered}
The implementation allows for the possibility that state
names are not instantiated (as required by the treatment of gaps). 
Moreover it uses memoization, and it
ensures that the predicate succeeds at most once:
\begin{numbered}\begin{verbatim}
connection(A,B):-
        (  var(A)          -> true
        ;  var(B)          -> true
        ;  A=:=B           -> true
        ;  B < A           -> fail   % word-graphs are acyclic
        ;  ok_conn(A,B)    -> true
        ;  fail_conn(A,B)  -> fail
        ;  wordgraph:trans(A,_,X,_),
           connection(X,B) -> assertz(ok_conn(A,B))
        ;  assertz(fail_conn(A,B)),
           fail 
        ).
\end{verbatim}\end{numbered}

A somewhat different approach that may turn out to be more efficient
is to use the ordinary comparison operator that we used in the
original definition of the head-corner parser.  The possible extra
cost of allowing impossible partial analyses is worthwhile if the
more precise check would be more expensive. If for typical input
word-graphs the number of transitions per state is high (such that
almost all pairs of states are connected), then this may be an
option.

\subsection{Accounting for Word-graph Scores}
In order to account for the acoustic score of a derivation (defined as
the sum of the acoustic scores associated with all transitions from the
word-graph involved in the derivation) we assume that the predicate
{\tt lexical\_analysis} represents the acoustic score of the piece of
the word-graph that it covers by an extra argument.  During the first
phase acoustic scores are ignored. During the second phase (when a
particular derivation is constructed) the acoustic scores are
combined. 

\section{Head-corner parsing and Robustness}
\label{robustness}
Certain approaches towards robust parsing use the partial results of
the parser. In such approaches it is assumed that even if no full
parse for the input could be constructed, the discovery of other
phrases in the input might still be useful. 
In order for such approaches to work it is often assumed
that a bottom-up parser is essential: parsers that use top-down
information (such as the head-corner parser)  
may fail to recognize relevant sub-parses in the
context of an ungrammaticality. 

In the application for which the head-corner parser was developed,
robust processing is essential. In a spoken dialogue system it is
often impossible to parse a full sentence, but in such cases the
recognition of e.g.\ temporal expressions might still be very
useful. Therefore, a robust processing technique which collects the
remnants of the parsing process in a meaningful way seems desirable. 

In this subsection we show how the head-corner parser can be used in
such circumstances. The approach consists of two parts. Firstly, the
parser is modified in such a way that it finds all derivations of the
start symbol {\em anywhere in the input}. Furthermore, the start
symbol should be defined in such a way that it includes all categories
which are considered useful for the application.

\subsection{Underspecification of the positions}

Normally the head-corner parser will
be called as e.g.\ :
\begin{verbatim}
?- parse(s(Sem),0,12).
\end{verbatim}
indicating that we want to parse a sentence from position 0 to 12 with
category {\tt s(Sem)} (a sentence with a semantic representation that
is yet to be discovered). Suppose however that a specific robustness
module is interested in all `maximal projections' anywhere in the
sentence. Such a maximal projection may be represented by a term
{\tt xp(Sem)}. Furthermore there may be unary grammar rules
rewriting such an {\tt xp} into appropriate categories, e.g.:
\begin{numbered}\begin{verbatim}
xp(Sem) --> np(Sem).  xp(Sem) --> s(Sem).
xp(Sem) --> pp(Sem).  xp(Sem) --> advp(Sem).
\end{verbatim}\end{numbered}
If we want to recognize all maximal projections at all positions in
the input, then we can simply give the following parse goal:
\begin{numbered}\begin{verbatim}
?- parse(xp(Sem),_,_).
\end{verbatim}\end{numbered}
Now one might expect that such an underspecified goal will dramatically
slow down the head-corner parser, but this turns out to be false. 
In actual fact we have experienced an increase of
efficiency using underspecification. This can only be understood in
the light of the use of memoization.
Even though we now have a much more general goal, the number
of different goals that we need to solve is much smaller.

Also note that even though the first call to the parse predicate has
variable extreme positions, this does not imply that all power of
top-down prediction is lost by this move; recursive calls to the parse
predicate may still have instantiated left and/or right extreme
positions.  The same applies with even more force  for top-down
information on categories. 

\subsection{The Robustness Component in OVIS}
In an attempt to obtain a robust natural language understanding
component we have experimented in OVIS with the techniques mentioned
in the preceding paragraph. The top category (start
symbol) of the OVIS grammar is defined as the category {\tt max(Sem)}.
Moreover there are unary rules such as {\tt max(Sem) $\rightarrow$
  np(Sem,..)} for {\sc np, s, pp, advp}.

In the first phase, the parser finds all occurrences of the top
category in the input word-graph. Thus, we obtain items for all
possible maximal projections anywhere in the input graph. In the
second phase, the robustness component selects a sequence of such
maximal projections.  The robustness procedure consists of a
best-first search from the beginning of the graph to the end of the
graph. A path in the input graph can be constructed by taking steps of
the following two types. To move from position $P$ to $Q$ you can
either:
\begin{itemize}
\item use a maximal projection from $P$ to $Q$ (as constructed by the
  parser)
\item use a transition from $P$ to $Q$. In this case we say that we
`skip' that transition. 
\end{itemize}

In order to compare paths in the best-first search method, we have
experimented with score functions which include some or all of the
following factors: 
\begin{itemize}
\item the number of skips. We prefer paths with a smaller number of such
  skips. 
\item the number of maximal projections. We prefer paths with a
  smaller number of such projections.
\item the combined acoustic score as defined in the word-graph.
\item the appropriateness of the semantic representation given the
  dialogue context
\item the bigram score.
\end{itemize}

If bigram scores are not included, then this best-first search
method can be implemented efficiently because for each state in the
word-graph we only have to keep track of the best path to
that state. 

The resulting `best' path in general consists of a number of maximal
projections. In the OVIS application these often are simple time or
place expressions. The pragmatic module is able to deal with such
unconnected pieces of information and will perform better if given
such partial parse results.

In order to evaluate the appropriate combination of the factors
determining the scoring function, and to evaluate this approach with
respect to other approaches, we use a corpus of word-graphs for which
we know the corresponding actual utterances.  We compare the sentence
associated with the `best' path in the word-graph with the sentence
that was actually spoken. Clearly if the robustness component more
often uses the information that was actually uttered, then we have more
confidence in that component. This notion of word accuracy is an
approximation of semantic accuracy (or `concept accuracy').  The string
comparison is defined by the minimal number of deletions and
insertions that is required to turn the first string into the second
(Levenshtein distance), although it may be worthwhile to investigate other
measures. For example, it seems likely that for our application it is
less problematic to `miss' information, whereas `hallucination' is a
more severe problem. This could be formalized by a scoring function in
which insertion (into analysis result) is cheaper than deletion.

Currently the best results are obtained with a scoring function in
which bigram scores, acoustic scores and the number of skips is
included.  We have also implemented a version of the system in which
acoustic scores and bigram scores are used to select the best path
through the word-graph. This path is then sent to the parser and the
robustness component. In this `best-1-mode' the system performs
somewhat worse in terms of word-accuracy, but much faster (cf.\ the
experiments in the next section). 

\section{Practical Experience}
\label{compar}

There does not exist a generally agreed upon method to measure the
efficiency of parsers for grammars of the kind we assume here, i.e.
constraint-based grammars for natural language understanding.
Therefore, I will present the results of the parser for the current
version of the OVIS grammar in comparison with a number of other
parsers that have been developed in the same project (by the author
and his colleagues).  Moreover, a similar experiment was performed with
two other grammars: the English MiMo2 grammar \cite{mimo2-article},
and the English Alvey NL Tools grammar \cite{anlt-grammar}.  It should be clear
that the results to be presented should not be taken as a formal
evaluation, but are presented solely to give an impression of the
practical feasibility of the parser, at least for its present purpose.
The following results should be understood with these reservations in
mind. 

\subsection{Other Parsers}

In the experiments the head-corner parser was compared with a number
of other parsers.  The parsers are described in further detail in
\namecite{ovis-deliverable-jan} and \namecite{ovis-deliverable-okt}.
The last two parsers of the following list were implemented by
Mark-Jan Nederhof. 

\begin{itemize}
\item {\tt lc}.  Left-corner parser. This parser is derived from the
  head-corner parser. It therefore uses many of the ideas presented
  above. Most importantly it uses selective memoization with goal
  weakening and packing. The parser is closely related to the BUP
  parser \cite{bup}.
\item {\tt bu-inactive}.  Inactive chart parser. This is a bottom-up
  parser which only records inactive edges. It uses packing. It
  uses a pre-compiled version of the grammar in which no empty
  productions are present.
\item {\tt bu-earley} Bottom-up Earley parser. This is a bottom-up
  chart parser which records both active and inactive items. It
  operates in two phases and uses packing. It uses a pre-compiled
  version of the grammar in which no empty productions are present.
\item {\tt bu-active} 
  Bottom-up Earley parser without packing. This is a chart parser
  which only constructs active items (except for categories which
  unify with the top category). It uses a pre-compiled
  version of the grammar in which no empty productions are present.
\item {\tt lr} LR parser. This is an experimental implementation
  of a generalization for Definite Clause Grammars of the parser
  described in \namecite{NE96}. It proceeds in a single phase and does
  not use packing. It uses a table to maintain partial analyses. It
  was not possible to perform all the experiments with this parser
  due to memory problems during the construction of the LR table. 
\end{itemize}

Note that we have experimented with a number of different versions of
each of these parsers. We will report only on the most efficient
version. The experiments were performed on a 125Mhz HP-UX 735 machine
with 240 Megabytes of memory. Timings measure CPU-time and should be
independent of the load on the machine. \footnote{Experiments suggest
  that the load on the machine in fact does influence the timings
  somewhat. However, the experiments were performed at times when the
  load of the machine was low. It is believed, therefore, that no such
  artifacts are present in the numbers below. }

\subsection{Experiment 1: OVIS}

The OVIS grammar (for Dutch) contains about 1400 lexical
entries (many of which are station and city names) and 66 rules (a
substantial fraction of those rules is concerned with time and date
expressions), including 7 epsilon rules. The most important
epsilon rule is part of a gap threading implementation of
verb-second. The grammar is documented in detail in
\shortcite{ovis-deliverable-jan}.
The head-corner table contains 128 pairs, the lexical
head-corner table contains 93 pairs, the gap-head-corner table
contains 14 pairs. The left-corner table contains 156 pairs, the
lexical left-corner table contains 114 pairs, the gap-left-corner
table contains 20 pairs. The pre-compiled grammar (which is used by
the chart parsers) contains 92 rules.

The input for the parser consists of a test-set of 5000 word-graphs,
randomly taken from a corpus of more than 25000 word-graphs.  These
word-graphs are the latest word-graphs that were available to us; they
are `real' output of the current version of the speech recognizer as
developed by our project partners. 
In this application, typical utterances are short. As a consequence,
the typical size of word-graphs is rather small too, as can be seen in
table~\ref{ltable}.

\begin{table}[h,t,b]
\begin{center}
\begin{tabular}{||r|r||}\hline
\# transitions & \# word-graphs\\\hline
  0-5   & 2825\\
  6-10  &  850\\
 11-15  &  408\\
 16-20  &  246\\
 21-30  &  237\\
 31-40  &  146\\
 41-50  &   83\\
 51-75  &  112\\
 76-100 &   44\\
101-150 &   36\\
151-200 &   12\\
263     &    1\\
\hline\end{tabular}
~~~
\begin{tabular}{||r|r||}\hline
\# words & \# utterances\\\hline
 1-2  & 2465\\
 3-4  & 1448\\
 5-6  &  543\\
 7-8  &  319\\
 9-10 &  118\\
11-12 &   56\\
13-14 &   26\\
15-16 &   20\\
17-18 &    5\\
\hline\end{tabular}
\end{center}
\caption{\label{ltable}The leftmost table gives information concerning
  the number of transitions per word-graph of the test set for the
  OVIS grammar. As can be seen from this table, more than half of the
  corpus consists of word-graphs with at most five transitions. In the
  rightmost table the number of words per utterance is given. Many
  utterances consists of less than five words. }
\end{table}

We report on three different experiments with the OVIS grammar and
these word-graphs. In the first experiment, the system runs in
best-1-mode: the best path is selected from the word-graph using
bigram scores and the acoustic scores (present in the word-graph).
This best path is then sent to the parser and robustness component. In
the second experiment, the parser is given the utterance as it was
actually spoken (to simulate a situation in which speech recognition
is perfect).  In the third experiment, the parser takes the full
word-graph as its input. The results are then passed on to the
robustness component. As explained in the previous section on
robustness, each of the parsers finds all derivations of the start symbol
anywhere in the input (this is the case in each of the OVIS
experiments). 

For the current version of the OVIS system, parsing on the basis of
the best path in the word-graph gives results in terms of
word-accuracy which are similar to the results obtained with full
word-graphs. Results for concept-accuracy are not yet available.
Details can be found in \namecite{ovis-deliverable-okt}.

\subsubsection{Parsing best path only}
In table~\ref{ovista} the CPU-time requirements and the
maximum space requirements of the
different parsers are listed.  
\begin{table}[h]
\begin{center}
\begin{tabular}{||l|r|r|r|r||}\hline
parser       & total (msec) & msec/sentence & maximum & maximum space\\ \hline
hc           & 169370       &   34          &    530  &         163  \\
lc           & 180160       &   36          &    530  &         171  \\
bu-active    & 291870       &   58          &   4220  &        1627  \\
bu-inactive  & 545060       &  109          &  13050  &         784  \\
bu-earley    & 961760       &  192          &  24470  &        2526  \\
lr           &1088940       &  218          & 416000  &        4412  \\
\hline\end{tabular}
\end{center}
\caption{\label{ovista}Total and average CPU-time and maximal space
  requirements for a test-set of 5000 best paths through word-graphs
  (OVIS grammar). }
\end{table}
In the table we list respectively the total number of milliseconds
CPU-time required for all 5000 word-graphs (timings include lexical
lookup, parsing and the robustness component), the average number of
milliseconds per word-graph, and the maximum number of milliseconds
for a word-graph. The final column lists the maximum amount of
space requirements (per word-graph, in Kbytes).  \footnote{These sizes are
obtained using the SICStus prolog built-in predicate {\tt
  statistics(program\_space,X)}. This only measures the size of
the internal database, but not the size of the stacks. The size of
stacks has never been a problem for any of the parsers; the size
of the internal database has occasionally led to problems for the
bottom-up chart parsers.}

\subsubsection{Parsing sentences}

The differences in CPU-time for the corpus of 5000 word-graphs are
similar to differences we have found for other test sets. The results are
also very similar to the results we obtain if we parse the actually
spoken utterances. Table~\ref{ovistb} lists the results
of parsing the set of 5000 utterances from which the word-graphs were
derived. 
\begin{table}[h]
\begin{center}
\begin{tabular}{||l|r|r|r|r||}\hline
parser       & total (msec) & msec/sentence & maximum & maximum space\\ \hline
hc           & 126930       &   25          &   510   &         137  \\
lc           & 137090       &   27          &   490   &         174  \\
bu-active    & 257390       &   51          &  4030   &        1438  \\
bu-inactive  & 546650       &  109          & 15170   &        1056  \\
bu-earley    & 934810       &  187          & 25490   &        3558  \\
lr           & 957980       &  192          &417580   &        4435  \\
\hline\end{tabular}
\end{center}
\caption{\label{ovistb}Total and average CPU-time and maximum 
space requirements for a test-set of 5000 utterances (OVIS grammar). }
\end{table}

\subsubsection{Parsing word-graphs}

Obviously, parsing word-graphs is more difficult than parsing only the
best path through a word-graph, or parsing an ordinary sentence. In
table~\ref{ovistc} we list the results for the same set of 5000
word-graphs. This experiment could only be performed for the
head-corner and the left-corner parser. The other parsers ran into
memory problems for some very large word-graphs.

\begin{table}[h]
\begin{center}
\begin{tabular}{||l|r|r|r|r||}\hline
parser       & total (msec) & msec/word-graph & maximum & maximum space\\\hline
lc           & 410670       &   82          &  15360  &      4455    \\
hc           & 435320       &   87          &  16230  &      4174    \\
\hline\end{tabular}
\end{center}
\caption{\label{ovistc}Total and average CPU-time and maximum 
space requirements for a test-set of 5000 word-graphs (OVIS grammar).}
\end{table}

In order to compare the other parsers too, I performed the experiment
with a time-out of 5000 msec (the memory problems only occur
for word-graphs that take longer to process).  In table~\ref{ovistd}
the percentage of word-graphs that can be treated within a certain
amount of CPU-time are listed.

\begin{table}[h]
\begin{center}
\begin{tabular}{||l|r|r|r|r|r|r|r||}\hline
parser      & 500  & 1000 & 2000 & 3000 & 4000 & 5000 & time-outs\\ \hline
lc          & 97.72 & 99.28 & 99.78 & 99.92 & 99.92 & 99.92 &   4\\
hc          & 97.42 & 98.94 & 99.60 & 99.84 & 99.92 & 99.92 &   4\\
lr          & 91.44 & 94.42 & 96.30 & 96.98 & 97.34 & 97.70 & 115\\ 
bu-active   & 91.84 & 94.76 & 96.04 & 96.84 & 97.30 & 97.60 & 120\\
bu-inactive & 82.36 & 88.64 & 92.24 & 94.10 & 95.14 & 95.86 & 207\\
bu-earley   & 77.10 & 84.26 & 89.04 & 91.42 & 92.64 & 93.50 & 325\\
\hline\end{tabular}
\end{center}
\caption{\label{ovistd}
  Percentage of word-graphs that can be treated within time
  limit (OVIS grammar). }
\end{table}

From the experiments with the OVIS grammar and corpus it can be
concluded that the head-corner and left-corner parsers (implemented
with selective memoization and goal-weakening) are much more efficient
than the other parsers. In the case of word-graphs, the left-corner
parser is about 5\% faster than the head-corner parser; for strings,
the head-corner parser is about 6 to 8\% faster than the left-corner
parser.

\subsection{Experiment 2: MiMo2}
Another experiment was carried out for the English grammar of the
MiMo2 system. This grammar is a unification-based grammar which is
compiled into a DCG. The grammar contains 525 lexical entries, 63
rules including 13 gaps. 
The head-corner relation contains 33 pairs and the lexical head-corner
relation contains 18 pairs.  The left-corner parser runs into hidden
left-recursion problems on the original grammar, so it uses a version
of the grammar in which left-most gaps are compiled out. This compiled
grammar has 69 rules. The left-corner relation contains 80 pairs; the
lexical left-corner relation contains 62 pairs. As a result, the
left-corner parser only hypothesizes gaps for non-left-most daughters.
Because the grammar never allows gaps as head, the head-corner parser
can be optimized in a similar fashion. Both the left-corner and
head-corner parser use a goal-weakening operator which only leave the
functor symbol. This simplifies the way in which the goal table is
maintained.

For this experiment we have no notion of typical input, but instead
just made up a set of 25 sentences of various length and difficulty,
with a total of 338 readings. In order to be able to complete the
experiment a time-out of 60 seconds of CPU-time was used. Timings
include lexical lookup and parse tree recovery.

The original parser implemented in the MiMo2 system (a left-corner
parser without packing) took 294 seconds of CPU-time
to complete the experiment (with 3 time-outs). Because the test
environment was (only slightly) different, we have indicated the latter
results in italics. Average CPU-time is only given for those
parsers which completed each of the sentences within the time
limit. The results are given in table~\ref{mimot}

\begin{table}[h]
\begin{center}
\begin{tabular}{||l|r|r|r|r||}\hline
parser       & total (msec) &msec/sentence &maximum space &time-outs\\ \hline
hc           &   52670      &  2107        &  2062        & 0\\
bu-active    &   52990      &  2120        & 30392        & 0\\
lc           &  109750      &  4390        &  8570        & 0\\
{\it mimo2-lc}&{\it 294000} &              &              & {\it 3}\\
bu-earley    &  439050      &              & 12910        & 4\\
bu-inactive  &  498610      &              &  7236        & 5\\
\hline\end{tabular}
\end{center}
\caption{\label{mimot}Total and average CPU-time and maximum 
space requirements for set of 25 sentences (MiMo2 grammar). }
\end{table}

The bottom-up active chart parser performs better on smaller sentences
with a small number of readings. For longer and more ambiguous
sentences the head-corner parser is (much) more efficient.  The other
parsers are consistently much less efficient.

\subsection{Experiment 3: Alvey NL Tools}
A final set of experiments was performed for the Alvey NL Tools
grammar \cite{anlt-grammar}, similar to the experiments discussed in
\namecite{carroll-anlt}. For a longer description of the grammar and
the test sets we refer to this publication. The grammar contains 2363
lexical entries, and 780 rules (8 of which are gaps). The left-corner
relation contains 440 pairs; the lexical left-corner relation contains
254 pairs. No gaps are possible as left-most element of the
right-hand-side of a rule.  

In order to be able to use the head-corner parser we needed to
determine for each of the rules which element on the
right-hand-side constitutes the head of the rule.  The head-corner
relation contains 352 pairs; the lexical head-corner relation contains
180 pairs. We also report on experiments in which for each rule the
left-most member of the right-hand-side was selected as the head.
The goal-weakening operator used for the left-corner and head-corner
parser removes all features (only leaving the functor symbol of each
category); again this simplifies the maintenance of
the goal table considerably. 

The bottom-up chart parsers use a version of the grammar in which all
epsilon rules are compiled out. The resulting grammar has 1015 rules.

The first test set consists of 129 short sentences (mean length 6.7
words). Our results were obtained with a newer version of the Alvey NL Tools
grammar. In the table below we list the results for the same grammar
and test set for Carroll's bottom-up left-corner parser
(BU-LC). Carroll performed this experiment on a SUN UltraSparc
1/140. It was estimated by Carroll and the author that this machine is
about 1.62 times faster than the HP-UX 735 on which the other
experiments were performed. \footnote{The SPECINT92 figures for the 
Ultra 1/140 and HP 735/125 confirm this: 215 and 136 respectively.}

In the table below we have multiplied the
13.3 seconds of CPU-time (obtained by Carroll) with this factor 
in order to compare his results with our results. Clearly, these
numbers should be taken with extreme caution, because many factors in the
test environment differ (hardware, LISP versus Prolog). For this
reason we use italics in table~\ref{anltta}.

\begin{table}[h]
\begin{center}
\begin{tabular}{||l|r|r|r||}\hline
parser        & msec        & msec/sentence & max Kbytes\\\hline
bu-active     & 18250       & 141           &  1276     \\
lc            & 21900       & 170           &   137     \\
{\it Carroll BU-LC} & {\it 21500} & {\it 167}     &           \\
hc (lc mode)  & 23690       & 184           &   165     \\
bu-earley     & 27670       & 214           &   758     \\
hc            & 68880       & 534           &   140     \\
bu-inactive   & 83690       & 649           &   170     \\
\hline\end{tabular}
\end{center}
\caption{\label{anltta}Total and average CPU-time and maximum 
space requirements for set of 129 short sentences (Alvey NL Tools
grammar). Italicized items are offered for cautious comparison. }
\end{table}


The second test set consists of 100 longer and much more complex
sentences. The length of the sentences is distributed uniformly
between 13 and 30 words (sentences created by Carroll). Many of the
sentences have many parses: the maximum number of parses is 2736 for
one 29-word sentence. Average number of readings is about 100 readings
per sentence. 

Again, we list the results Carroll obtained with the BU-LC parser.  It
took 
205.7 seconds on the SUN UltraSparc 1/140. 
\footnote{Note that Carroll reports on recognition times only, whereas
  our results include the construction of all individual parse trees.
  For this experiment the left-corner parser used about 163 seconds on
  recognition. However, in the recognition phase the parser ignores a
  number of syntactic features and therefore this number cannot be
  compared fairly with Carroll's number either.} 
The bottom-up active chart
parser ran into memory problems for some very ambiguous sentences and
was very slow on many of the other sentences (due to the lack of
packing). The results are summarized in table~\ref{anlttb}.

\begin{table}[h]
\begin{center}
\begin{tabular}{||l|r|r|r||}\hline
parser           & msec        & msec/sentence& max Kbytes\\\hline
lc            &  195850      & 1959           & 10955     \\
hc (lc mode)  &  216180      & 2162           & 10969     \\
{\it Carroll BU-LC} & {\it 333000} & {\it 3330}    &           \\
bu-earley     & 1219120      &12191           & 18232     \\
hc            & 3053910      &30539           &  7915     \\
bu-inactive   & 3578370      &35784           & 16936     \\
bu-active     & $>>$   &                & $> 65000$\\
\hline\end{tabular}
\end{center}
\caption{\label{anlttb}Total and average CPU-time and maximum 
space requirements for set of 100 longer sentences (Alvey NL Tools grammar).
Italicized items are offered for cautious comparison.  }
\end{table}


The implementation of the left-corner parser based on selective
memoization and goal-weakening seems to be substantially more
efficient than the chart-based implementation of Carroll. The
head-corner parser running in left-corner mode is almost as fast as
this specialized left-corner parser. This suggests that the use of
selective memoization with goal-weakening is on the right track.  

From these experiments it can be concluded that the head-corner parser
is not suitable for the Alvey NL Tools grammar. The reason seems to be that for
this grammar the amount of top-down information that is available
through the head-corner table is of limited value. In order to parse a
given goal category, too many different lexical head-corners are
typically available. For example, in order to parse a sentence
possible head-corners include auxiliaries, verbs, adverbs,
complementizers, pronouns, prepositions, determiners, nouns and
conjunctions. In contrast, in the MiMo2 grammar only verbs can
function as the head-corners of sentences.  As a result the prediction
step introduces too much non-determinism. A related reason for the
poor performance for this grammar might be the large amount of lexical
ambiguity. The grammar and lexicon used in the experiment is compiled
from a compact user notation. In the compiled format, all disjunctions
are spelled out in different rules and lexical entries.  As a result,
many words have a large number of (only slightly different) readings.
It may be that the head-corner parser is less suitable in such
circumstances. This could also explain the fact that the head-corner
parser performs better on strings then on word-graphs: in many
respects the generalization to word-graphs is similar to an increase
in lexical ambiguity. This suggests that further improvements in the
design of the head-corner parser should be sought in the prediction
step. 

\subsection*{Availability test material}
  The material used to perform the experiments with the MiMo2 grammar
  and the Alvey NL Tools grammar, including several versions of the head-corner
  parser, is available via anonymous ftp and the
  world-wide-web. The material is ready to be plugged into the Hdrug
  environment available from the same site.

\begin{center}
\tt
ftp://ftp.let.rug.nl/pub/prolog-app/CL97/ \\
http://www.let.rug.nl/\verb+~+vannoord/CL97/
\end{center}

\starttwocolumn
\begin{acknowledgments} 
  Some of the introductory material of this article is a modified and
  extended version of the introduction of \namecite{bouma-gertjan}.
  Gosse Bouma, John Carroll, Rob Koeling, Mark-Jan Nederhof, John
  Nerbonne and three anonymous ACL reviewers provided useful feedback.
  The grammar used in the OVIS experiments is written by Gosse Bouma,
  Rob Koeling and myself. Mark-Jan Nederhof implemented the bottom-up
  active chart parser and the experimental LR parser.  The MiMo2
  grammar is written by Joke Dorrepaal, Pim van der Eijk and the
  author. I am very grateful to John Carroll for his help in making
  the experiments with the Alvey NL Tools grammar possible.  This
  research was carried out within the framework of the Priority
  Programme Language and Speech Technology (TST). The TST-Programme is
  sponsored by NWO (Dutch Organization for Scientific Research).
\end{acknowledgments}

\bibliographystyle{fullname}

\begin{thebibliography}{}

\bibitem[\protect\citename{Alshawi}1992]{cle-book}
Alshawi, Hiyan, editor.
\newblock 1992.
\newblock {\em The Core Language Engine}.
\newblock ACL-MIT press.

\bibitem[\protect\citename{Billot and Lang}1989]{billot-lang}
Billot, S. and B.~Lang.
\newblock 1989.
\newblock The structure of shared parse forests in ambiguous parsing.
\newblock In {\em 27th Annual Meeting of the Association for Computational
  Linguistics}, pages 143--151, Vancouver.

\bibitem[\protect\citename{Bouma and van Noord}1993]{bouma-gertjan}
Bouma, Gosse and Gertjan van Noord.
\newblock 1993.
\newblock Head-driven parsing for lexicalist grammars: Experimental results.
\newblock In {\em Sixth Conference of the European Chapter of the Association
  for Computational Linguistics}, Utrecht.
\newblock Available from http://www.let.rug.nl/\verb+~+vannoord/papers/.

\bibitem[\protect\citename{Boves \bgroup et al.\egroup }1995]{ovisplan}
Boves, Lou, Jan Landsbergen, Remko Scha, and Gertjan van Noord.
\newblock 1995.
\newblock {\em Language and Speech Technology}.
\newblock NWO Den Haag.
\newblock Project plan for the NWO Priority Programme `Language and Speech
  Technology'. Available from http://grid.let.rug.nl:4321/.

\bibitem[\protect\citename{Carroll}1994]{carroll-anlt}
Carroll, John.
\newblock 1994.
\newblock Relating complexity to practical performance in parsing with
  wide-coverage unification grammars.
\newblock In {\em 32th Annual Meeting of the Association for Computational
  Linguistics}, pages 287--294, Las Cruces, New Mexico.

\bibitem[\protect\citename{Grover, Carroll, and Briscoe}1993]{anlt-grammar}
Grover, Claire, John Carroll, and Ted Briscoe.
\newblock 1993.
\newblock The alvey natural language tools grammar (4th release).
\newblock Technical Report 284, Computer Laboratory, Cambridge University UK.

\bibitem[\protect\citename{Johnson and D\"orre}1995]{johnson-doerre}
Johnson, Mark and Jochen D\"orre.
\newblock 1995.
\newblock Memoization of coroutined constraints.
\newblock In {\em 33th Annual Meeting of the Association for Computational
  Linguistics}, pages 100--107, Boston.

\bibitem[\protect\citename{Kay}1989]{kay-hd}
Kay, Martin.
\newblock 1989.
\newblock Head driven parsing.
\newblock In {\em Proceedings of Workshop on Parsing Technologies}, Pittsburg.

\bibitem[\protect\citename{Lang}1989]{lang-atr}
Lang, Bernard.
\newblock 1989.
\newblock A generative view of ill-formed input processing.
\newblock In {\em ATR Symposium on Basic Research for Telephone Interpretation
  (ASTI)}, Kyoto Japan.

\bibitem[\protect\citename{Lavelli and Satta}1991]{satta-berlin}
Lavelli, Alberto and Giorgio Satta.
\newblock 1991.
\newblock Bidirectional parsing of lexicalized tree adjoining grammar.
\newblock In {\em Fifth Conference of the European Chapter of the Association
  for Computational Linguistics}, Berlin.

\bibitem[\protect\citename{Matsumoto \bgroup et al.\egroup }1983]{bup}
Matsumoto, Y., H.~Tanaka, H.~Hirakawa, H.~Miyoshi, and H.~Yasukawa.
\newblock 1983.
\newblock {BUP}: a bottom up parser embedded in {Prolog}.
\newblock {\em New Generation Computing}, 1(2).

\bibitem[\protect\citename{Nederhof and Satta}1994]{markjan-satta}
Nederhof, Mark-Jan and Giorgio Satta.
\newblock 1994.
\newblock An extended theory of head-driven parsing.
\newblock In {\em 32th Annual Meeting of the Association for Computational
  Linguistics}, New Mexico State University.

\bibitem[\protect\citename{Nederhof and Satta}1996]{NE96}
Nederhof, M.J. and G.~Satta.
\newblock 1996.
\newblock Efficient tabular {LR} parsing.
\newblock In {\em 34th Annual Meeting of the Association for Computational
  Linguistics}, pages 239--246, Santa Cruz.

\bibitem[\protect\citename{van Noord}1991]{head-corner}
van Noord, Gertjan.
\newblock 1991.
\newblock Head corner parsing for discontinuous constituency.
\newblock In {\em 29th Annual Meeting of the Association for Computational
  Linguistics}, Berkeley.
\newblock Available from http://www.let.rug.nl/\verb+~+vannoord/papers/.

\bibitem[\protect\citename{van Noord}1993]{vannoord-diss}
van Noord, Gertjan.
\newblock 1993.
\newblock {\em Reversibility in Natural Language Processing}.
\newblock {Ph.D.} thesis, University of Utrecht.
\newblock Available from http://www.let.rug.nl/\verb+~+vannoord/papers/.

\bibitem[\protect\citename{van Noord}1994]{vannoord-tag}
van Noord, Gertjan.
\newblock 1994.
\newblock Head corner parsing for {TAG}.
\newblock {\em Computational Intelligence}, 10(4).
\newblock Available from http://www.let.rug.nl/\verb+~+vannoord/papers/.

\bibitem[\protect\citename{van Noord}1995]{acl95}
van Noord, Gertjan.
\newblock 1995.
\newblock The intersection of finite state automata and definite clause
  grammars.
\newblock In {\em 33th Annual Meeting of the Association for Computational
  Linguistics}, MIT Boston.
\newblock Available from http://www.let.rug.nl/\verb+~+vannoord/papers/.

\bibitem[\protect\citename{van Noord \bgroup et al.\egroup
  }1996]{ovis-deliverable-okt}
van Noord, Gertjan, Gosse Bouma, Rob Koeling, and Mark-Jan Nederhof.
\newblock 1996.
\newblock Conventional natural language processing in the nwo priority
  programme on language and speech technology. {October 1996 Deliverables}.
\newblock Technical Report~28, NWO Priority Programme Language and Speech
  Technology.
\newblock http://grid.let.rug.nl:4321/.

\bibitem[\protect\citename{van Noord \bgroup et al.\egroup
  }1991]{mimo2-article}
van Noord, Gertjan, Joke Dorrepaal, Pim van~der Eijk, Maria Florenza, Herbert
  Ruessink, and Louis des Tombe.
\newblock 1991.
\newblock An overview of {MiMo2}.
\newblock {\em Machine Translation}, 6:201--214.
\newblock Available from http://www.let.rug.nl/\verb+~+vannoord/papers/.

\bibitem[\protect\citename{van Noord \bgroup et al.\egroup
  }1996]{ovis-deliverable-jan}
van Noord, Gertjan, Mark-Jan Nederhof, Rob Koeling, and Gosse Bouma.
\newblock 1996.
\newblock Conventional natural language processing in the nwo priority
  programme on language and speech technology. {January 1996 Deliverables }.
\newblock Technical Report~22, NWO Priority Programme Language and Speech
  Technology.
\newblock http://grid.let.rug.nl:4321/.

\bibitem[\protect\citename{Pereira and Shieber}1987]{pereira-shieber}
Pereira, Fernando~C.N. and Stuart~M. Shieber.
\newblock 1987.
\newblock {\em {Prolog} and Natural Language Analysis}.
\newblock Center for the Study of Language and Information Stanford.

\bibitem[\protect\citename{Pereira and Warren}1980]{dcg}
Pereira, Fernando~C.N. and David Warren.
\newblock 1980.
\newblock Definite clause grammars for language analysis - a survey of the
  formalism and a comparison with augmented transition networks.
\newblock {\em Artificial Intelligence}, 13.

\bibitem[\protect\citename{Rosenkrantz and Lewis-II}1970]{lc}
Rosenkrantz, D.J. and P.M. Lewis-II.
\newblock 1970.
\newblock Deterministic left corner parsing.
\newblock In {\em IEEE Conference of the 11th Annual Symposium on Switching and
  Automata Theory}, pages 139--152.

\bibitem[\protect\citename{Sahlin}1991]{mixtus}
Sahlin, Dan.
\newblock 1991.
\newblock {\em An Automatic Partial Evaluator for Full Prolog}.
\newblock {Ph.D.} thesis, The Royal Institute of Technology (KTH) Stockholm,
  Sweden.
\newblock SICS Dissertation Series 04.

\bibitem[\protect\citename{Satta and Stock}1989]{satta-stock}
Satta, Giorgio and Oliviero Stock.
\newblock 1989.
\newblock Head-driven bidirectional parsing. {A tabular method}.
\newblock In {\em Proceedings of the Workshop on Parsing Technologies}, pages
  43--51, Pittsburg.

\bibitem[\protect\citename{Shieber}1985]{shieber-restriction}
Shieber, Stuart~M.
\newblock 1985.
\newblock Using restriction to extend parsing algorithms for
  complex-feature-based formalisms.
\newblock In {\em 23th Annual Meeting of the Association for Computational
  Linguistics}, Chicago.

\bibitem[\protect\citename{Shieber \bgroup et al.\egroup }1990]{cl}
Shieber, Stuart~M., Gertjan van Noord, Robert~C. Moore, and Fernando~C.N.
  Pereira.
\newblock 1990.
\newblock Semantic-head-driven generation.
\newblock {\em Computational Linguistics}, 16(1).
\newblock Available from http://www.let.rug.nl/\verb+~+vannoord/papers/.

\bibitem[\protect\citename{Sikkel}1993]{sikkel-diss}
Sikkel, Klaas.
\newblock 1993.
\newblock {\em Parsing Schemata}.
\newblock {Ph.D.} thesis, Twente University, Enschede.
\newblock Published in 1997 by Springer Verlag, Texts in Theoretical Computer
  Science, An EATCS Series.

\bibitem[\protect\citename{Sikkel and op~den Akker}1992]{sikkel}
Sikkel, Klaas and Rieks op~den Akker.
\newblock 1992.
\newblock Head-corner chart parsing.
\newblock In {\em Proceedings Computer Science in the Netherlands (CSN '92)},
  Utrecht.

\bibitem[\protect\citename{Sikkel and op~den Akker}1993]{sikkel-durbuy}
Sikkel, Klaas and Rieks op~den Akker.
\newblock 1993.
\newblock Predictive head-corner chart parsing.
\newblock In {\em IWPT 3, Third International Workshop on Parsing
  Technologies}, pages 267--276, Tilburg/Durbuy.

\bibitem[\protect\citename{Tomita}1987]{tomita}
Tomita, M.
\newblock 1987.
\newblock An efficient augmented context-free parsing algorithm.
\newblock {\em Computational Linguistics}, 13(1-2):31--46.

\bibitem[\protect\citename{Warren}1992]{ds-warren}
Warren, David~S.
\newblock 1992.
\newblock Memoing for logic programs.
\newblock {\em Communications of the ACM}, 35(3):94--111.

\end{thebibliography}

\end{document}